\documentclass[10pt,twocolumn,letterpaper]{article}

\usepackage[pagenumbers]{iccv} % To force page numbers, e.g. for an arXiv version

\usepackage{multirow}
\usepackage{makecell}
\usepackage{pifont}
\newcommand{\cmark}{\ding{51}}%
\newcommand{\xmark}{\ding{55}}%
\usepackage{xcolor,colortbl}  % to highlight rows in table

\usepackage{tikz}

\usepackage[accsupp]{axessibility}  % Improves PDF readability for those with disabilities.

\definecolor{paleYellow}{RGB}{252,249,186}
\definecolor{aubergine}{RGB}{38,14,70}
\definecolor{verylightgrey}{RGB}{221,221,221}

\DeclareRobustCommand{\pvol}{synthetic occupancy field\xspace}

\DeclareRobustCommand{\pvolabbr}{SOF\xspace}

\DeclareRobustCommand{\refm}{view completion model\xspace}
\DeclareRobustCommand{\REFM}{View Completion Model\xspace}

\DeclareRobustCommand{\Refmabbr}{VCM\xspace}

\DeclareRobustCommand{\recon}{scene reconstruction model\xspace}
\DeclareRobustCommand{\RECON}{Scene Reconstruction Model\xspace}

\DeclareRobustCommand{\Reconabbr}{SRM\xspace}

\definecolor{iccvblue}{rgb}{0.21,0.49,0.74}
\usepackage[pagebackref,breaklinks,colorlinks,allcolors=iccvblue]{hyperref}

\def\paperID{4691} % *** Enter the Paper ID here
\def\confName{ICCV}
\def\confYear{2025}

\title{Dream-to-Recon: Monocular 3D Reconstruction with Diffusion-Depth Distillation from Single Images}
\author{
Philipp Wulff$^{1}$ \qquad
Felix Wimbauer$^{1,2}$ \qquad
Dominik Muhle$^{1,2,3}$ \qquad
Daniel Cremers$^{1,2}$ 
\vspace{0.3em} \\
{\normalsize $^1$Technical University of Munich} \quad
{\normalsize $^2$MCML} \quad
{\normalsize $^3$SE3 Labs} 
\vspace{0.42em}
\\
{\tt\normalsize \href{https://philippwulff.github.io/dream-to-recon}{philippwulff.github.io/dream-to-recon}}
}

\begin{document}
% \maketitle

\twocolumn[{%
\renewcommand\twocolumn[1][]{#1}%
\maketitle
\begin{center}
    \setlength{\abovecaptionskip}{1pt}
    \centering
    \captionsetup{type=figure}
    \includegraphics[width=\textwidth,trim=0 .0cm 0 0, clip]{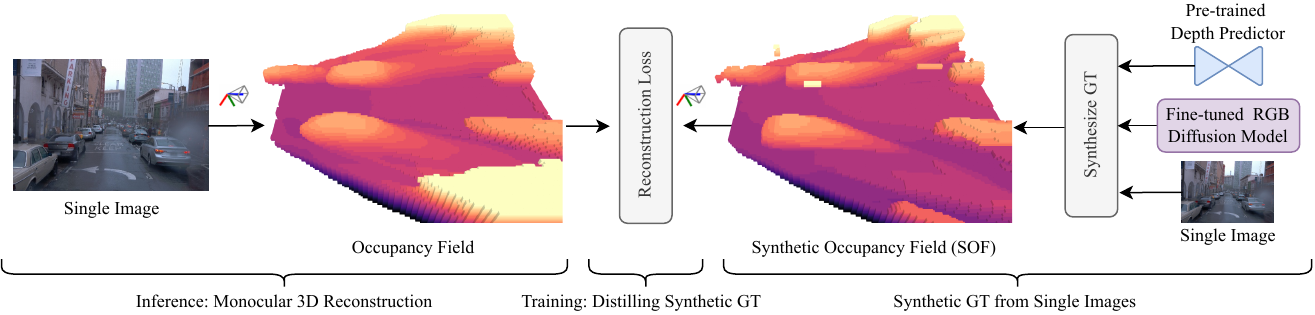}
    \captionof{figure}{\textbf{Dream-to-Recon.} We leverage fine-tuned diffusion models for inpainting and a pre-trained depth predictor to generate high-quality scene geometry from a single image, as seen on the right side. Afterwards, we distill a feed-forward \recon, which performs on par with reconstruction methods trained with multi-view supervision. The color in the plots indicates the up-coordinate.
    % Left: Show supervision with single input image (synthetic views built from it). Right: simple architecture overview; show inputs (input image), then simplified model, then output 3d reconstruction
    }
\end{center}%
}]

\begin{abstract}
Volumetric scene reconstruction from a single image is crucial for a broad range of applications like autonomous driving and robotics. 
Recent volumetric reconstruction methods achieve impressive results, but generally require expensive 3D ground truth or multi-view supervision. 
We propose to leverage pre-trained 2D diffusion models and depth prediction models to generate synthetic scene geometry from a single image. 
This can then be used to distill a feed-forward scene reconstruction model.
Our experiments on the challenging KITTI-360 and Waymo datasets demonstrate that our method matches or outperforms state-of-the-art baselines that use multi-view supervision, and offers unique advantages, for example regarding dynamic scenes. For more details and code, please check out our  \href{https://philippwulff.github.io/dream-to-recon}{project page}.
\end{abstract}
    
\section{Introduction}
\label{sec:intro}

% Monocular 3D reconstruction models are trained with multi-view data due to lack of real-world 3D ground truth. Multi-view data is still more expensive to collect than single images. Ither methods make use of subsequent frames in monocular video, but dynamic objects are a problem there.

% - some methods have shown that single-scene NeRF can be improved via learned priors on the optimization... but these were used for per-scene reconstruction and rely on NeRF-style training data with large baselines between the views. Therefore, these methods have not been demonstrated on large-scale outdoor scenes.

% Our approach trains a learned prior on single-image data. It uses this and a pre-trained depth predictor for monocular 3d reconstruction from synthetic novel views. Thus, we do not need multi-view or monocular video data. Further, we show that distilling these synthetic novel views into a zero-shot monocular 3d reconstructor improves robustness. We show competitiveness of our method against baselines trained with multi-view supervision (BTS, KYN). In summary, our contributions are:

Understanding the 3D geometry of a scene from a single image is fundamental for a wide range of applications, in particular for autonomous driving and robotics. 
A dense reconstruction of the environment enables machines to react to their surroundings and to reason about further actions such as path planning.

Traditionally, this task has been mainly approached via monocular depth estimation (MDE), where a network predicts per-pixel distance values for a given single image.
Through architectural improvements, scaled-up datasets with high-quality ground-truth, and cross-dataset training, these methods have achieved strong generalization capabilities and can be used in an off-the-shelf way \cite{ranftl2020towards, bhat2023zoedepth, yin2023metric3d, piccinelli2024unidepth}.
In particular, MDE methods are now even able to predict metric-scale depth maps.
While MDE is already useful for many applications, it has a severe limitation:
Depth maps are only a 2D projection of the 3D scene, and do not capture the scene's full 3D structure.
Consequently, it is not possible to infer information about areas that are beyond the parts visible in the input image.
This limitation makes pure MDE unsuitable for many 3D understanding tasks, e.g. planning the path of a vehicle into a parking spot that was only partially observed.

To address this shortcoming, a new line of work aims to infer a \textit{volumetric 3D reconstruction} of the scene from a single image \cite{wimbauer2023behind, li2024know, cao2022monoscene, li2023voxformer, pan2024renderocc, sima2023scene}.
Here, the 3D geometry is either represented as a discrete voxel grid \cite{cao2022monoscene, li2023voxformer, sima2023scene} or as a continuous density field \cite{wimbauer2023behind, li2024know, pan2024renderocc}.
Many approaches are trained in a fully-supervised way using dense 3D annotations.
However, such 3D ground truth is difficult and expensive to obtain, e.g. by accumulating Lidar scans from a moving vehicle over time.
Even then, dynamic scenes with many moving objects pose a significant challenge, as accumulation over time can lead to trailing artifacts and inconsistencies.
Recent works propose to avoid the need for explicit 3D ground-truth from Lidar by using a self-supervised loss based on multi-view consistency and volume rendering \cite{wimbauer2023behind, li2024know, huang2024selfocc, pan2024renderocc}.
Notably, Behind the Scenes (BTS) \cite{wimbauer2023behind} demonstrates accurate volumetric reconstruction, even in occluded areas, by relying on multi-view image data captured from a moving car.
Despite its impressive performance, BTS still requires a special multi-camera setup and accumulation of views over time to have sufficient observations of the scene.
Furthermore, this training scheme is still not robust towards dynamic objects.

Recently, there has also been impressive progress in the field of scene generation, enabled by the success of image diffusion models.
Several works combine image diffusion models with pre-trained depth predictors to generate novel views of a scene using the \textit{render-refine-repeat} framework \cite{hollein2023text2room, chung2023luciddreamer, li2022infinitenature}.
Starting from an input image, they first warp the pixels into a virtual novel view using the depth predictor.
Then, they inpaint and refine the warped image using an image diffusion model, which is conditioned on the warped pixels.
By repeating this process multiple times, it is possible to generate several consistent views of a 3D environment and subsequently optimize a 3D representation such as 3D Gaussian Splatting.
So far, these methods have focused on generating high-fidelity novel views of constrained scenes. 
However, the generated geometry, which is important for many downstream tasks, is still lacking in quality.
Furthermore, these approaches are very compute-intensive and not (yet) suitable for real-time applications.

In this work, we propose a specialized approach based on the \textit{render-refine-repeat} framework to generate high-quality geometry reconstructions of a scene from a single image.
% \todo{@Philipp add 1-2 sentences with details}
Given a color image and its predicted depth, we render novel views, jointly remove artifacts and inpaint occluded regions.
Subsequently, we fuse all views into a 3D occupancy field.
These generated scene geometries are then distilled into a robust feed-forward model for single-image volumetric 3D reconstruction.
The distillation involves both 3D volumetric supervision and 2D rendering supervision from virtual novel views.
To deal with inherent ambiguities in occluded regions, our model also predicts 3D uncertainty.
During both scene generation and distillation, we exclusively rely on single image and do not require multi-view data or 3D supervision.

Experiments on the challenging KITTI-360 \cite{liao2022kitti360} and Waymo \cite{sun2020waymo} datasets show that our distilled model is competitive with state-of-the-art methods for monocular 3D reconstruction using multi-view supervision.
Furthermore, we show that our method has unique advantages when it comes to dynamic scenes.
Through extensive ablation studies, we validate the effectiveness of our design choices both for scene generation and distillation.
Our contributions are:

\begin{itemize}
    % \item depth gradient based occlusion detection with density based 3d representation
    \item A specialized \textit{\refm} that inpaints and refines synthetic novel views and which can be trained using only a single image per scene.
    \item A \textit{\pvol} formulation to construct dense scene geometry from multiple synthetic views.
    % \item method for synthesizing 3d consistent novel views with information about occlusions using a RGB diffusion models and pretrained depth predictor 
    % - depth gradient based occlusion detection with density based 3d representation
    \item A single-view \textit{\recon}, which is trained from the synthesized data and is competitive with reconstruction methods trained on multi-view data.
\end{itemize}
\section{Related Work}
\label{sec:related_work}

\subsection{Monocular depth prediction}

The task of inferring depth information from a single image is a fundamental challenge in computer vision.
% Over the years, there have emerged several distinct lines of works:
Early works trained neural networks in a supervised way to directly regress relative depth for specific domains, \eg autonomous driving \cite{eigen2014depth, eigen2015predicting, wang2015towards, liu2015learning, laina2016deeper, fu2018deep, guizilini2021sparse}. 
Some works avoided the need for explicit supervision and relied on self-supervised loss formulations based on multi-view consistency \cite{godard2017unsupervised, godard2019digging}.
As one of the first, Midas \cite{ranftl2020towards} proposed a training formulation that allows effective training on many diverse datasets at the same time, shifting the focus to build more general-purpose monocular relative depth prediction networks. 
A number of follow-up approaches have since improved different parts of the pipeline \cite{ranftl2021vision, Wei2021CVPR, bhat2023zoedepth, yang2024depth, yang2025depth}.
Most recently, several works \cite{hu2024metric3d, yin2023metric3d, piccinelli2024unidepth, bochkovskii2024depth}, notably Metric3D~\cite{hu2024metric3d} and UniDepth~\cite{piccinelli2024unidepth}, proposed methods for general-purpose \textit{metric} depth prediction.
As monocular depth prediction methods have matured over the years, many applications now use such models in an off-the-shelf fashion.

\subsection{Scene as occupancy}

While depth predictions are useful, they only capture 2.5D information about the scene and do not provide details in e.g. occluded regions.
Recently, a new line of work started training neural networks to directly infer a volumetric reconstruction from a single image \cite{wimbauer2023behind, cao2022monoscene, cao2023scenerf}.
Volumetric reconstructions allow to determine binary occupancy values for any point in the observed camera frustum.
In general, the scene is either represented as a dense voxel grid \cite{cao2022monoscene, li2023voxformer, miao2023occdepth, li2023fb, sima2023scene}, a continuous density field \cite{wimbauer2023behind, hayler2024s4c, cao2023scenerf, pan2024renderocc, li2024know}, or hybrid approaches like tri-planes \cite{huang2023tri, huang2024selfocc}.
To obtain ground truth data for supervision, many works accumulate Lidar scans from a moving sensor over time \cite{li2023voxformer, miao2023occdepth, li2023fb, sima2023scene, huang2023tri}.
Other approaches use losses on multi-view supervision and photometric consistency and can thus avoid the need for explicit 3D data \cite{wimbauer2023behind, hayler2024s4c, cao2023scenerf, pan2024renderocc, huang2024selfocc, li2024know}.

In general, methods for volumetric reconstruction from a single image require complex sensor setups to generate training data.
This limits the scale of these datasets and makes generalization challenging.
Furthermore, they struggle with dynamic objects which violate the static scene assumption. 
Many approaches either ignore \cite{wimbauer2023behind} or filter out \cite{pan2024renderocc} dynamic objects at training time.
In contrast, our approach can be trained using single images and can deal with dynamic objects by design.
It only requires an off-the-shelf diffusion model and depth prediction model.

\subsection{3D scene generation}

Following the recent success of diffusion models for image generation \cite{rombach2022high, podell2023sdxl, ramesh2021zero, saharia2022photorealistic, chen2023pixart}, researchers also aimed to create powerful 3D generative models.
However, similar to 3D reconstruction, the amount of training data available for 3D generation is limited. 
Consequently, models that directly generate 3D are restricted to object-centric content \cite{hong2023lrm, wei2024meshlrm}.

Several lines of work propose to leverage 2D image generation models to improve 3D generation.
% As these models are trained on massive datasets, they develop strong priors over the composition of realistic images.
% Through different techniques, this pior can be leveraged for 3D content generation.
Inspired by DreamBooth~\cite{poole2022dreamfusion}, some works \cite{wang2023prolificdreamer, qian2023magic123, shi2023mvdream} apply score distillation sampling (SDS) to generate 3D objects.
However, such models are not stable and are thus still limited to object-centric 3D generation.
Another promising direction \cite{wei2024meshlrm, tang2024lgm, gao2024cat3d, xu2024grm, hollein2024viewdiff, karnewar2023holodiffusion, szymanowicz2023viewset, liu2023zero, siddiqui2025meta} finetunes or retrains 2D diffusion models to generate multiple, 3D consistent views of a scene from different viewing directions.
The generated views then enable 3D reconstruction in a second step.
While these approaches often benefit from the generalization capabilities of pretrained 2D diffusion models, they are inherently limited.
Besides requiring multi-view training data, the views that the models can generate are predefined and usually arranged in a circular pattern around an object in the center. 
Therefore, this line of work also struggles with unbounded in-the-wild scenes.
Finally, a number of methods \cite{liu2021infinite, li2022infinitenature, hollein2023text2room, schult2024controlroom3d, chung2023luciddreamer, shriram2024realmdreamer, engstler2024invisible, zhou2024dreamscene360} follow an iterative paradigm called \textit{render-refine-repeat}.
Every iteration is made up of two steps: 1) The current scene is \textit{rendered} into an unobserved, virtual view, 2) the virtual view is \textit{refined} (inpainted) using a 2D generative model and then lifted back into the scene using a pretrained depth prediction model.
Earlier works like Infinite Nature~\cite{liu2021infinite} and Text2Room \cite{hollein2023text2room} represent the scene as 3D meshes, while more recent approaches use NeRFs~\cite{zhang2024text2nerf} or Gaussian Splatting~\cite{chung2023luciddreamer, shriram2024realmdreamer}.
Unlike this work, most approaches in that domain focus on novel-view-synthesis, which does not require perfect geometry.
% \todo{Talk about the challenges in this paradigm}
% \todo{put into context with our method. explain why ours is different}

% In combination with techniques such as Score Distillation Sampling (SDS)~\cite{poole2022dreamfusion} or ControlNet~\cite{zhang2023adding}, they also enable applications in 3D generation.

\begin{figure*}[t]
    \centering
    \includegraphics[width=\linewidth, trim={0 0cm 0 0cm},clip]{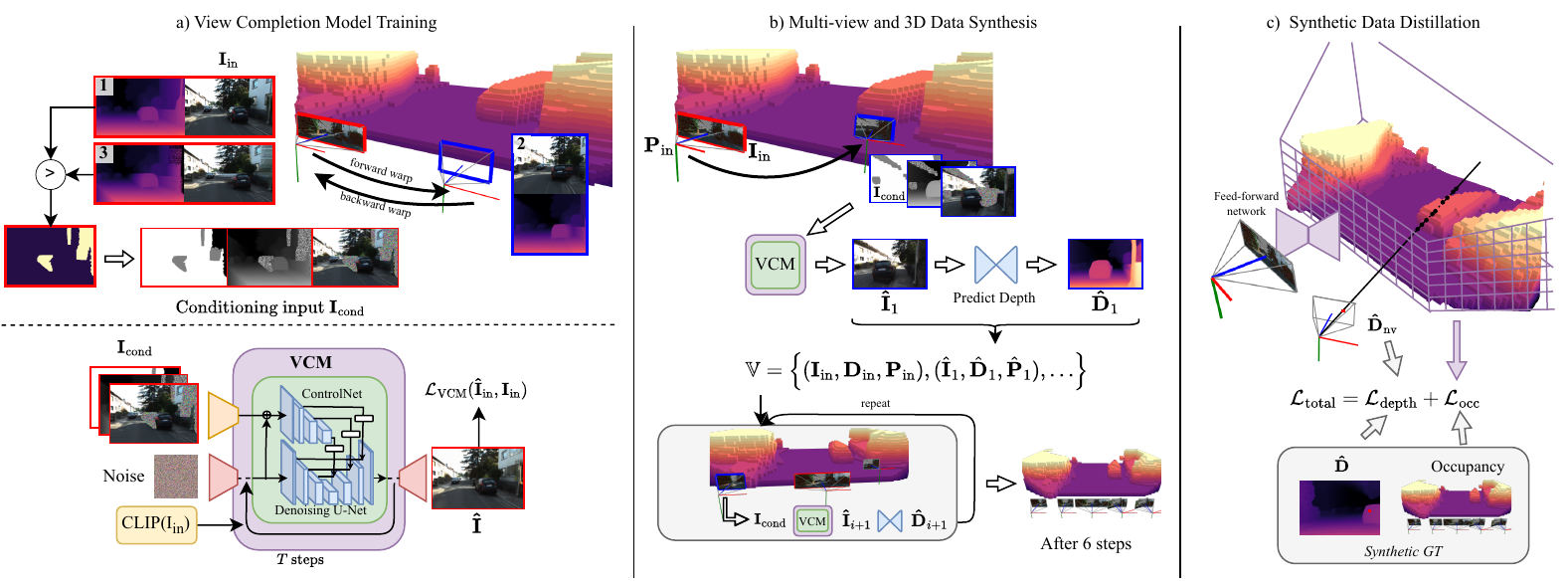}
    \caption{\textbf{Method overview.} \textbf{a)} We train a view completion model (VCM) that inpaints occluded areas and refines warped views. Training uses only a single view per scene and leverages forward-backward warping for data generation. \textbf{b)} The VCM is applied iteratively alongside a depth prediction network to synthesize virtual novel views, enabling progressive refinement of the 3D geometry. \textbf{c)} The synthesized scene geometries are then used to distill a feed-forward \recon. by supervising occupancy and virtual depth.
    % \todo{col a: VCM: show training setup in 3D and conditioning image input to model in the input view + another novel to show that it generalizes | col b: encoding with input, then synthetic view gen, then encoding with all views | col c: distillation of synthetic data into feed-forward model}\todo{@philipp change loss names accordingly}
    }
    \label{fig:method}
    \vspace{-.5cm}
\end{figure*}

\section{Method}
\label{sec:method}

% We detail the construction of a 3D density and radiance field from multiple RGBD images, which we term \textit{\pvol} (\pvolabbr) (\cref{sec:pseudo_volume}). We describe our method for training a \refm\ on single images (\cref{sec:corrupted_training_data_gen}) and for synthesizing multi-view and 3D ground truth (\cref{sec:synth_gt}).

In the following, we first introduce our \textit{\refm} (\Refmabbr), which completes occlusions and removes artifacts in warped images.
We then present how we apply the \Refmabbr in combination with a depth prediction network to iteratively synthesize 3D scene geometry in the form of a \textit{\pvol} (\pvolabbr) from just a single image.
Finally, we describe how we distill these synthesized scenes into a lightweight feed-forward \textit{\recon}.
\cref{fig:method} provides a comprehensive overview.

\subsection{Preliminaries}
\label{ssec:preliminaries}

For a given scene, our method receives as input a single image $\mathbf{I}_{\text{in}} \in ([0, 1]^3)^\Omega$, where  $\Omega = \{1, \ldots, H\} \times \{1, \ldots, W\}$ denotes the grid of pixels. 
Through an off-the-shelf monocular depth prediction network% $\mathcal{D}(\mathbf{I}): ([0, 1]^3)^\Omega \rightarrow ([0, \infty ))^\Omega$
, we can additionally obtain per-pixel depths $\mathbf{D}_\mathbf{I} \in \mathbb{R}_+^\Omega$.
For a synthetic novel view $\hat{\mathbf{I}}_i$, let $\mathbf{P}_{i} \in \operatorname{SE3}$ be the camera pose and let $\mathbf{P}_{i \rightarrow j}$ denote a relative camera pose between views $i$ and $j$. 
For all views, we assume the same projection $\pi(\mathbf{x}): \mathbb{R}^3 \rightarrow \mathbb{R}^2$ and unprojection $\pi^{-1}(\mathbf{p}, d): (\mathbb{R}^2, \mathbb{R}_+) \rightarrow \mathbb{R}^3$ function, which map from 3D points $\mathbf{x} \in \mathbb{R}^3$ to pixels $\mathbf{p} \in \mathbb{R}^2$ with depth $d_\mathbf{p} \in \mathbb{R}_+$, and vice-versa.
By combining $\pi$ and $\mathbf{P}$, we can map a pixel from the input view into another view:
\begin{equation}
    \mathbf{p}_i = \pi\left(\mathbf{P}_{\text{in}\rightarrow i} \pi^{-1}(\mathbf{p}_\text{in}, d_{\mathbf{p}_\text{in}})\right)
\end{equation}

\subsection{Training the \REFM}
\label{sec:corrupted_training_data_gen}

A single image only provides a partial observation of the 3D scene.
We aim to use the ability of 2D generative models to inpaint realistic content given some context.
Concretely, we follow the \textit{render-refine-repeat} framework.
First, the known scene content (\eg the input image) is warped into a virtual novel view, where unknown regions remain empty.
Then, a generative model fills in the empty or occluded regions conditioned on the known context and ideally removes artifacts.
The now completed and refined view can be lifted into 3D again to refine the scene.
However, naively applying an inpainting diffusion model like StableDiffusion-2.1 yields insufficient inpainting quality.
This is because the views found in \eg autonomous driving differ significantly from the ones the model was originally trained on, which tend to be casual photos.
Therefore, we rely on a specialized \textit{\refm} (\Refmabbr).

The \Refmabbr is implemented as a ControlNet \cite{zhang2023controlnet} for a pre-trained StableDiffusion-2.1-unCLIP diffusion model. 
Rich conditioning and the correct parameterization are very important for effective view completion.
Thus, the \Refmabbr receives as input the warped RGB, the (incomplete) depth representing the current scene geometry, and an inpainting mask. 
We mask out areas for inpainting with random noise in all inputs and apply morphological closing to the mask. 
Additionally, we provide the model with the CLIP \cite{radford2021clip} embedding of the input image.
From this input, the \Refmabbr produces an inpainted and refined image.

The \Refmabbr needs to be trained using ground-truth pairs of conditioning and target image.
Since we have only a single image per scene available, we take inspiration from \cite{cai2022diffdreamer} and design a training scheme based on forward-backward warping.
Given an input image $\mathbf{I}_{\text{in}}$ and predicted depth $\textbf{D}_{\mathbf{I}_{\text{in}}}$, we first warp the pixels into a virtual novel view with a random camera pose.
Then, the virtual view is warped back into the input camera pose.
For areas of the input image, which would be occluded in the novel view, the backward warped image will be corrupted.
We can find these occluded regions by doing a per-pixel comparison of the back-projected depth with the original depth.
If the reprojected depth is lower for a pixel, then that pixel is occluded.
Through this process, we create realistic occlusions and rendering artifacts  to train the \Refmabbr.
The warping uses the same rendering formulation as for synthesizing a new scene which is described in the following.
% \todo{add equations, add figure references}
% the re-projected depth for those regions will lie in front of the original input image depth. 
% It jointly refines and inpaintsof corrupted novel views. We generate the training image pairs from single images, inspired by \cite{cai2022diffdreamer}. Given an RGBD image, we create a training pair by warping the image to a novel view camera pose via the \pvol\ and then re-projecting it to the original pose, hence \cref{fig:corrupted_training_data_gen}. If the input image shows regions in 3D space that are occluded in the novel view, the re-projected depth for those regions will lie in front of the original input image depth. Thus, we keep the re-projected image as the corrupted image, the input image as GT and obtain an inpainting mask from the comparison of the depths.
% \begin{figure}
%     \centering
%     \includegraphics[width=\linewidth]{figures/gen_training_data.png}
%     \caption{Caption}
%     \label{fig:corrupted_training_data_gen}
% \end{figure}

\subsection{Synthesizing Scene Geometry}
\label{ssec:pseudo_volume}

% \todo{we want to represent our scene in 3D using an IOF via input imgs + potential synthetic views and use them to do volume rendering}
% \todo{shorten B equations}
Within the \textit{render-refine-repeat} framework, we generate several synthetic novel views, complete them using the \Refmabbr, predict corresponding depth maps, and update the scene geometry.
To this end, we require a way to model scene geometry from the predicted depth maps, render novel views, and determine occlusions.
Let $\mathbb{V} = \left\{(\textbf{I}_\text{in}, \textbf{D}_\text{in}, \textbf{P}_\text{in}), (\hat{\textbf{I}}_i, \hat{\textbf{D}}_i, \hat{\textbf{P}}_{i}), \ldots\right\}$ be the set of currently active view-depth-pose triplets.
Initially, $\mathbb{V}$ contains only the input view and is then extended via \textit{render-refine-repeat}.
% We also have to determine 

\paragraph{Modeling scene geometry.}
Throughout our approach, we consider a continuous \textit{\pvol} $\Theta_\mathbb{V}(\textbf{x}): \mathbb{R}^3\rightarrow \{0, 1\}$, which maps every point $\textbf{x} \in \mathbb{R}^3$ in the scene to a binary occupancy value and which depends on $\mathbb{V}$.
Intuitively, a point $\textbf{x}$ is regarded as not occupied if it lies \textit{in front} of the predicted depth of \textit{any} active view, for which it is in the camera frustum.
Conversely, if it lies behind the predicted depth for \textit{all} views, it is considered occupied.
Let $\textbf{x}_i = \textbf{P}_i \textbf{x}$ be the point transformed into the coordinate system of view $i$, $\textbf{p}_i = \pi(\textbf{x}_i)$ the corresponding pixel after projection to the image plane, and $d_{\textbf{p}_i}$ the predicted depth at that pixel.
We determine the occupancy of $\textbf{x}$ through the following logical formula:

\begin{gather}
    \theta^\text{inside-frustum}_i(\textbf{x}) = \left[\textbf{p}_i \in \Omega\right]\\
    \theta^\text{behind-depth}_i(\textbf{x}) = \left[\textbf{x}_i^z > d_{\textbf{p}_i}\right]\\
    \theta^\text{occ}_i(\textbf{x}) = \theta^\text{inside-frustum}_i(\textbf{x}) \land \theta^\text{behind-depth}_i(\textbf{x})\\
    \Theta_\mathbb{V}(\textbf{x}) = \bigwedge_{i \in \mathbb{V}} \theta_i^\text{occ}(\textbf{x})
\end{gather}

\paragraph{Rendering novel views.}  
We rely on the established volume rendering formulation \cite{kajiya1984ray}, popularized in recent years by NeRF \cite{mildenhall2021nerf}.
Here, an image is rendered by casting a ray for every pixel from the camera into 3D space.
The ray is evaluated at discrete steps $\mathbf{x}$ for density $\sigma(\mathbf{x})$ and color $c(\mathbf{x})$.
The final pixel color is determined by integrating over the density along the ray.

In our case, we can directly use our binary occupancy as density: $\sigma(\mathbf{x}) = \Theta_\mathbb{V}(\mathbf{x})$.
Because color information is not defined in 3D space, we sample it from the 2D views.
When projecting $\mathbf{x}$ into view $i$ and obtaining pixel location $\mathbf{p}_i$, let $\mathbf{c}_i$ be the corresponding pixel color.
However, not all $\mathbf{c}_i$ are valid, as the 3D point could be occluded or in empty space in the image.
To aggregate the sampled color values, we use a heuristic: 
If $\mathbf{x}$ is closer than a threshold $\tau_s$ to a surface in view $i$, then it is likely valid.
The final color is obtained by averaging all valid color values:
\begin{gather}
    \theta_i^\text{surface}(\mathbf{x}) = \left[\left|\textbf{x}_i^z - d_{\textbf{p}_i}\right| < \tau_s\right] \land \theta^\text{inside-frustum}_i(\textbf{x})\\
    c(\mathbf{x}) = \frac{\sum_{i\in \mathbb{V}} \mathbf{c}_i \cdot \theta_i^\text{surface}(\mathbf{x})}{\sum_{i\in \mathbb{V}} \theta_i^\text{surface}(\mathbf{x})}
\end{gather}

Let $\tilde{\textbf{I}}_\text{nv}$ denote the rendered view, which can contain invalid regions (due to occlusions) and sampling artifacts.
By computing the expected ray termination depth (which in our case corresponds to where a ray hits occupancy for the first time), we can also render a depth map $\tilde{\textbf{D}}_\text{nv}$.

\paragraph{Detecting occlusions in novel views.} 
When warping one view into another and the viewpoint change is limited, then sharp changes in the reference depth map serve as a good proxy for which regions will introduce occlusions.
We leverage this aspect and compute a depth gradient map using Sobel filters on the inverse depth map, which highlights regions with significant depth changes.
Subsequently, a threshold $\tau_d$ is applied to binarize the map.
\begin{align}
    \textbf{O}_i = \sqrt{\left(\frac{\partial}{\partial x}\textbf{D}_i\right)^2 + \left(\frac{\partial}{\partial y}\textbf{D}_i\right)^2} > \tau_d
\end{align}
We finally render it alongside the RGB color channels into the novel view to obtain $\tilde{\textbf{O}}_\text{nv}$.
In a post-processing step, we apply morphological opening to remove isolated noise followed by morphological closing join neighboring patches.

\paragraph{Sampling and completing novel views.}
By adding more views to the set of active views $\mathbb{V}$, we refine the scene geometry represented by the \pvol.
The better the views cover different areas of the scene, the better the resulting geometry will be.
Therefore, we predefine a set $\mathbb{P} = \{\textbf{P}_1, \ldots, \textbf{P}_{n_\text{synth}}\}$ of camera poses for synthetic novel views.
For every view $j \in \mathbb{P}$, we render the RGB $\tilde{\textbf{I}}_j$, depth $ \tilde{\textbf{D}}_j$, and occlusions $\tilde{\textbf{O}}_j$.
From this conditioning, the \Refmabbr then inpaints and refines the incomplete view, resulting in the synthetic novel view $\hat{\textbf{I}}_j$.
Finally, we apply the depth predictor to obtain the completed depth map $\hat{\textbf{D}}_j$ and we can update $\mathbb{V}$ with the triplet $\left(\hat{\textbf{I}}_j, \hat{\textbf{D}}_j, \hat{\textbf{P}}_j\right)$.

\subsection{Distilling into a \RECON}

While we can synthesize high-quality scene geometry using the \refm, the process is expensive and not suitable for real-time applications.
Furthermore, the pipeline involves several independent components, which makes it prone to artifacts.
To address these limitations, we generate scene geometry for all images of a dataset and distill a feed-forward \textit{\recon}.
Given a single image as input, it predicts a discretized occupancy field $\Theta_{\text{\Reconabbr}}\in [0, 1]^{Z \times H \times W}$.
However, there is an inherent ambiguity when generating or reconstructing scenes from a single image.
Therefore, we also predict an uncertainty field $\omega\in \mathbb{R}_+^{Z \times H \times W}$.
% to predict the 3D density of all content inside of the frustum given a single input image and supervise it via our synthetic 3D and multi-view data. Formally, we learn the mapping $\Gamma: \mathbf{I}\rightarrow\{\sigma_\mathbf{x},u_\mathbf{x}\vert \mathbf{x}\in frustum\}$, where $u_\mathbf{x}$ is the predicted uncertainty.

The training involves two loss terms. 
First, we directly supervise the predicted occupancy.
Concretely, we rely on an uncertainty-based loss formulation inspired by \cite{kendall2017uncertainties}, which is self-calibrating.
% This allows the model to downweight the regions of the 3D geometry, in which the ground truth
\begin{align}
    \mathcal{L}_{\text{occ}}(\Theta_{\text{\Reconabbr}}, \Theta) = \sum_\mathbf{x}\frac{(\Theta(\mathbf{x}) - \Theta_{\text{\Reconabbr}}(\mathbf{x}))^2}{\omega(\mathbf{x})} + \log\omega(\mathbf{x})
\end{align}
Additionally, we render depth maps $\tilde{\textbf{D}}_i$ using our $\Theta_{\text{\Reconabbr}}$ from the same camera poses as the synthetic views in $\mathbb{V}$.
We then directly supervise them via the depth predictions $\textbf{D}_i$ from $\mathbb{V}$ using a Gaussian Negative Log-Likelihood (GNLL) loss, as proposed in \cite{roessle2022dense}.
The loss term provides training signals to the surface areas of the predicted density field, which are particularly hard to learn.
\begin{equation}
    \mathcal{L}_\text{depth} = \sum_{i\in \mathbb{V}} \operatorname{GNLL}(\tilde{\textbf{D}}_i, \textbf{D}_i)
\end{equation}
The final loss is then a weighted sum:
\begin{equation}
    \mathcal{L} = \lambda_\text{occ} \mathcal{L}_{\text{occ}} + \lambda_\text{depth} \mathcal{L}_{\text{depth}}.
\end{equation}

\section{Experiments}
\label{sec:experiments}

% Our evaluation procedure systematically evaluates each component of our pipeline as follows: First we show novel view synthesis results for a simple and the final versions of our image refinement model. Then we use the full model to evaluate our occlusion detection strategy, novel view camera sampling and pseudo volume configuration. This leads towards producing optimal synthetic 3D GT. Finally, we show results of the distillation of synthetic data into a zero-shot reconstructor in ablations and compare against baselines.

To demonstrate the capabilities of our approach, we conduct extensive experiments regarding volumetric scene reconstruction both for the synthesized scenes and the distilled reconstruction model.
Furthermore, we carefully validate our design choices for the \refm, scene synthesis, and distillation procedure.
\begin{table*}[t]
    \centering
    \small
\begin{tabular}{lccccc|ccc}
    \toprule
    & \multirow{2}{*}[-0.16cm]{\makecell{Learns\\occlusions}} & \multirow{2}{*}[-0.16cm]{\makecell{No multi-\\view data}} & \multicolumn{3}{c}{KITTI-360 \cite{liao2022kitti360}} & \multicolumn{3}{c}{Waymo \cite{sun2020waymo}} \\
    \cmidrule(lr){4-6} \cmidrule(lr){7-9}
    \textit{Method} &  &  & $\text{O}_\text{acc}$ ↑ & $\text{IE}_\text{acc}$ ↑ & $\text{IE}_\text{rec}$ ↑ & $\text{O}_\text{acc}$ ↑ & $\text{IE}_\text{acc}$ ↑ & $\text{IE}_\text{rec}$ ↑ \\
    \midrule
    $\text{UniDepth}^\star$ \cite{piccinelli2024unidepth} & \xmark & \cmark & 0.89 & -- & -- & \underline{0.96} & -- & -- \\
    $\text{Metric3D-ViT-Large}^\star$ \cite{yin2023metric3d} & \xmark & \cmark & 0.91 & -- & -- & \textbf{0.97} & -- & -- \\
    $\text{BTS}^\star$ \cite{wimbauer2023behind} & \cmark & \xmark & \underline{0.92} & 0.69 & 0.64 & 0.95 & 0.63 & \underline{0.94} \\
    BTS + depth supervision & \cmark & \xmark & \underline{0.92} & 0.70 & 0.66 & 0.95 & 0.61 & \textbf{0.96} \\
    $\text{KYN}^\dagger$ \cite{li2024know} & \cmark & \xmark & \underline{0.92} & 0.70 & \underline{0.72}  & -- & -- & -- \\
    \midrule
    \textbf{Ours} & \cmark & \cmark & \textbf{0.93} & \textbf{0.72} & \textbf{0.75}  & \textbf{0.97} & \textbf{0.73} & \textbf{0.96} \\
    \textbf{Ours (Distilled)} & \cmark & \cmark & 0.90 & \underline{0.71} & 0.71  & \underline{0.96} & \underline{0.72} & 0.93 \\
    \bottomrule
\end{tabular}

        % scene_O_acc: 0.951398736834526
        % scene_IE_acc: 0.6100749060971772
        % scene_IE_rec: 0.9614664997583554

        % scene_O_acc: 0.9539574860930443
        % scene_IE_acc: 0.6327642391994596
        % scene_IE_rec: 0.9437476765594365

        % scene_O_acc: 0.9589493615031243
        % scene_IE_acc: 0.719843492358923
        % scene_IE_rec: 0.9297775938830993

        % scene_O_acc: 0.9298232369492406
        % scene_IE_acc: 0.720488907321069
        % scene_IE_rec: 0.7488819378434689

        % scene_O_acc: 0.951398736834526
        % scene_IE_acc: 0.6100749060971772
        % scene_IE_rec: 0.9614664997583554

        % scene_O_acc: 0.9017535901457205
        % scene_IE_acc: 0.7052711366771538
        % scene_IE_rec: 0.706006281731032

% 40 sweeps:
% ours:
        % scene_O_acc: 0.9610560801029205
        % scene_IE_acc: 0.6478166517242789
        % scene_IE_rec: 0.737311874806881

% BTS D:
        % scene_O_acc: 0.9558154556751252
        % scene_IE_acc: 0.5478946187514812
        % scene_IE_rec: 0.7753373437821865

% 300 sweeps:
% ours:
        % scene_O_acc: 0.9613345174193382
        % scene_IE_acc: 0.6492498921975494
        % scene_IE_rec: 0.7359426615014673

% BTS D:
        % scene_O_acc: 0.9562154556512833
        % scene_IE_acc: 0.55070004815422
        % scene_IE_rec: 0.7709151250720024
    \vspace{-.3cm}
    \caption{\textbf{Quantitative scene reconstruction.} We report cene geometry estimation results using ground-truth derived from accumulated LiDAR scans and semantic annotations on KITTI-360 and Waymo. Both the synthesized scenes and the distilled model match or surpass baselines, despite not requiring multi-view training data. \textbf{Legend}: $\star$: official checkpoint, $\dagger$: results as reported in the reference.}
    \label{tab:baselines_quantitative}
    \vspace{-.3cm}
\end{table*}

\subsection{Setup}

\paragraph{Data.}
We test our method on the challenging KITTI-360 \cite{liao2022kitti360} and Waymo \cite{sun2020waymo} self-driving datasets. 
Both datasets contain scenes with complex layouts and possibly dynamic objects.
While they provide multi-view data, we use only a single view per scene both for training and testing. 
For KITTI-360, we load images at $384\times1280$ resolution during \REFM training and at $192\times640$ resolution otherwise. 
On Waymo, we use images at a resolution of $320\times480$. 
% When training the monocular reconstructor, we combine the images from the front, front-left and front-right cameras during GT synthesis and only feed the front-camera's image to the monocular reconstructor model. 

\paragraph{Implementation details.}
% For our \refm, we rely on Stable-Diffusion-2.1 \cite{rombach2022high} and the standard ControlNet architecture for conditioning. 
% During training data synthesis, we render novel views at $192\times 360$ resolution, then re-render the input views at $192\times 288$ resolution and upscale to $512\times768$, since this is close to the resolution during Stable-Diffusion's training. 
% The input image is produced from the re-rendered image. 
% We use DDIM sampling \cite{todo} during training and evaluate using UniPCMultiStep sampling \cite{todo} with 5 steps. 
% We initialize from the official stable diffusion weights and then train for 20 epochs with batch size 20 and a constant learning rate of $10^{-5}$. 
% For the depth predictor, we use UniDepth \cite{piccinelli2024unidepth} during ControlNet training, 
For the depth predictor, we rely on Metric3D \cite{yin2023metric3d}. 
% We render the pseudo-volume using 48 coarse, 16 fine and 16 ray samples placed within $\sigma=2$ of the depth estimate.
The \recon adopts the ResNet50-based U-Net backbone from \cite{wimbauer2023behind}. 
We add batch-normalization layers to the backbone's decoder to stabilize mixed-precision training. 
% The backbone predicts a frustum-aligned $192\times 640\times c$ density grid with $c\in\{64, 128\}$ and inverse growth in the $z$ coordinate. 
The predicted occupancy field has a resolution $Z=64$, inversely spaced between $3m$ and $50m$.
% We first generate a \pvol for every image in the dataset, and then train the feed-forward model for 30 epochs. %with 64 coarse ray samples, batch size 48 and learning rate $10^{-3}$.
The training takes approximately 2 days for the \REFM and 3 days for the reconstruction model on 4 Nvidia A100 GPUs.
For more details regarding data setup and training, please refer to the supplementary material.
% \paragraph{Efficient reconstructor training.} UniPCMultiStepScheduler, AMP, Caching and batch size increase on epoch 2 \todo{put in appendix}

\paragraph{Evaluation procedure.}
We use 2D reconstruction metrics to evaluate image synthesis results in the input camera's viewpoint, where ground truth is available.
Here, PSNR measures reconstruction quality of RGB images, and absolute relative error (Abs Rel) is used for depth.
% \begin{align}
%     \text{PSNR}(\mathbf{I}_\text{pred}, \mathbf{I}_\text{GT}) &= 10\cdot\log_{10}{\frac{\text{MAX}^2}{\text{MSE}(\mathbf{I}_\text{pred}, \mathbf{I}_\text{GT})}} 
%     \\
%     \text{Abs-Rel}(\textbf{D}_\text{pred}, \textbf{D}_\text{GT}) &= \frac{1}{\vert M\vert}\sum_{i\in M}\frac{\vert D_{\text{pred}, i} - D_{\text{GT}, i} \vert}{D_{\text{GT}, i}}
% \end{align}
%
% where $\text{MSE}(\mathbf{I}_\text{pred}, \mathbf{I}_\text{GT})$ is the mean squared error and $\text{MAX}$ represents the maximum possible value of an image pixel. 
% We apply 2D reconstruction metrics to evaluate the input view as well as the novel view quality. 
% Evaluation in the input view involves applying the LDM4R refinement in the input view and resembles the LDM4R training setup.
When evaluating novel view quality, we apply refinement in the novel view and then re-project the refined views into the input view to compute the metrics w.r.t. the input image (masking out invalid pixels).
% This second evaluation serves to quantify the quality of synthetic GT.

Furthermore, we rely on the established occupancy evaluation strategy using LiDAR GT from \cite{wimbauer2023behind,li2024know} to evaluate 3D consistency of novel views and the quality of the 3D reconstruction in an area of dimension $x=[-4m,4m],\ y=[-1m,0m],\ z=[4m,20m]$.
On Waymo, we augment the LiDAR GT with occupancy from object bounding boxes to keep dynamic objects.
We evaluate the overall accuracy of the reconstruction $\text{O}_\text{acc}$, as well as the accuracy and recall of the reconstruction in invisible and empty regions (\ie behind the visible surface in the input image).
% This is particularly interesting, as it tests the quality of the reconstructions \textit{beyond} what a MDE model could achieve.
% 
% For any timestamp, a 3D occupancy grid is constructed by carving away all voxels that contain any points from 300 consecutive LiDAR sweeps. Subsequently, we evaluate the overall accuracy of the reconstruction $\text{O}_\text{acc}$ in the cuboid spanning $x=[-4m,4m],\ y=[-1m,0m],\ z=[4m,20m]$ in the input camera coordinate system. Because dynamic objects are more prevalent in the Waymo datset, we accumulate 20 LiDAR sweeps and keep occupancy within 3D GT bounding boxes. For Waymo, we report metrics on the validation set, since 3D bounding boxes not available on the test set.
% In addition, the metrics $\text{IE}_\text{acc}$ and $\text{IE}_\text{rec}$ describe the accuracy and recall for the reconstruction of .
% This is particularly interesting, as it tests the quality of the reconstructions \textit{beyond} what a MDE model could achieve.

\subsection{Scene Reconstruction}
We first test our method's ability to generate accurate volumetric geometry from a single image in complex scenes.
Here, the state-of-the-art volumetric reconstruction methods Behind the Scenes (BTS) \cite{wimbauer2023behind} and Know Your Neighbor (KYN) \cite{li2024know} serve as baselines.
Both use multi-view supervision from all cameras and multiple time steps. 
Because our approach uses an off-the-shelf depth predictor to synthesize geometry, we also train a second BTS baseline that uses depth prediction in addition to multi-view supervision.
This ensures similar priors and a fair comparison.
Furthermore, we test the recent MDE models UniDepth \cite{piccinelli2024unidepth} and Metric3D \cite{hu2024metric3d}.
Since depth prediction cannot reason about occluded areas, we do not report the $\text{IE}_\text{acc}$ and $\text{IE}_\text{rec}$ metrics. 

As shown in \cref{tab:baselines_quantitative}, compared to previous methods, our reconstructed \pvol achieve the best overall reconstruction ($\text{O}_\text{acc}$) and the best reconstruction of invisible and empty regions ($\text{IE}_\text{acc}$ and $\text{IE}_\text{rec}$). 
Our distilled \recon matches the performance of KYN and surpasses BTS \cite{wimbauer2023behind}, despite using only single images.
Even with depth supervision, BTS performance still lags behind.
We hypothesize that this lack of improvement stems from the strong depth cues already inherent in multi-view data.
In fact, inconsistent depth predictions with slightly varying scales may even harm BTS training.
% \todo{Add some sentences about Waymo}

\begin{figure*}[t]
    \centering
    \setlength{\tabcolsep}{2pt}
    \begin{tabular}{m{.03\textwidth} >{\centering\arraybackslash}m{.23\textwidth} >{\centering\arraybackslash}m{.23\textwidth} > {\centering\arraybackslash}m{.23\textwidth} >{\centering\arraybackslash}m{.23\textwidth}}
        \rotatebox{90}{\small Input} & 
        \includegraphics[width=\linewidth]{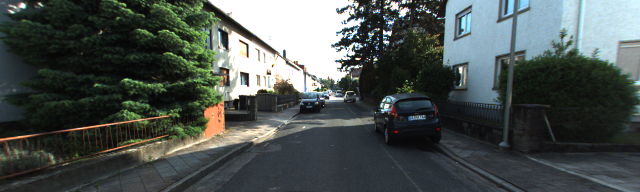} & 
        \includegraphics[width=\linewidth]{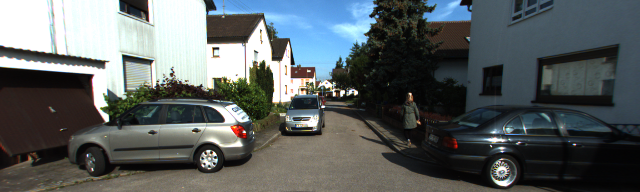} & 
        \raisebox{\height}{
            \tikz[baseline]{
              \node[inner sep=0pt,
                    fill=verylightgrey,
                    minimum width=\linewidth,
                    minimum height=.3\linewidth] 
                    (bg) at (0,0) {};
              \node[inner sep=0pt] 
                    at (bg.center) 
                    {\includegraphics[width=.45\linewidth]{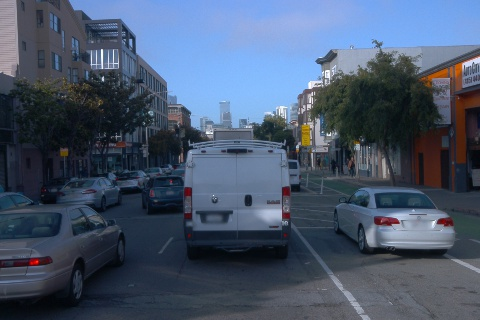}};
            }
        }
        &
        \raisebox{\height}{
            \tikz[baseline]{
              \node[inner sep=0pt,
                    fill=verylightgrey,
                    minimum width=\linewidth,
                    minimum height=.3\linewidth] 
                    (bg) at (0,0) {};
              \node[inner sep=0pt] 
                    at (bg.center) 
                    {\includegraphics[width=.45\linewidth]{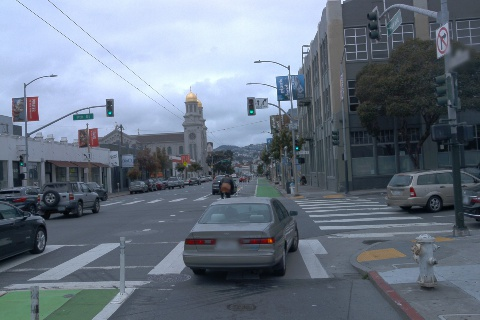}};
            }
        }
  \\

        \rotatebox{90}{\small Metric3D \cite{hu2024metric3d}} & 
        \includegraphics[width=\linewidth, trim=0cm 0cm 0cm 3cm, clip]{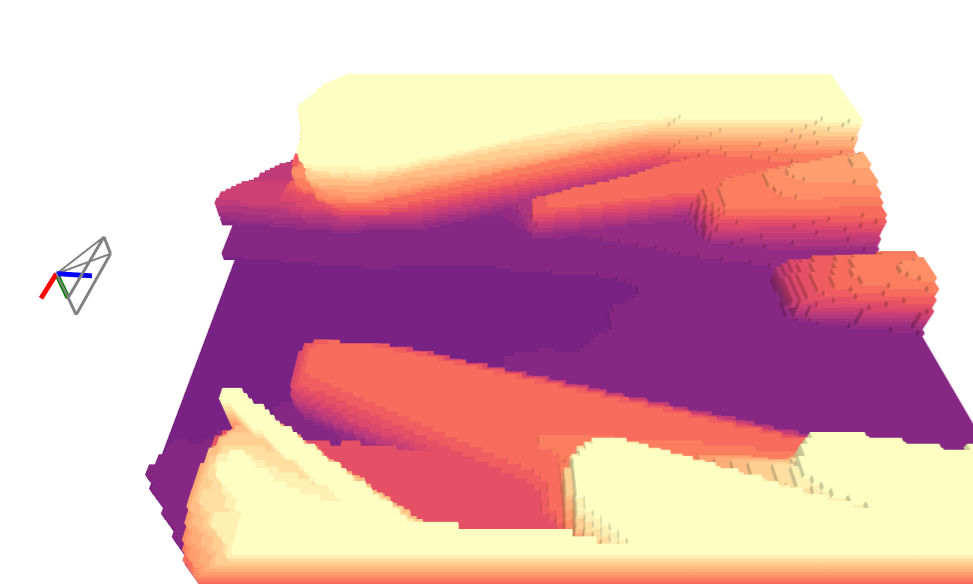} & 
        \includegraphics[width=\linewidth, trim=0cm 0cm 0cm 3cm, clip]{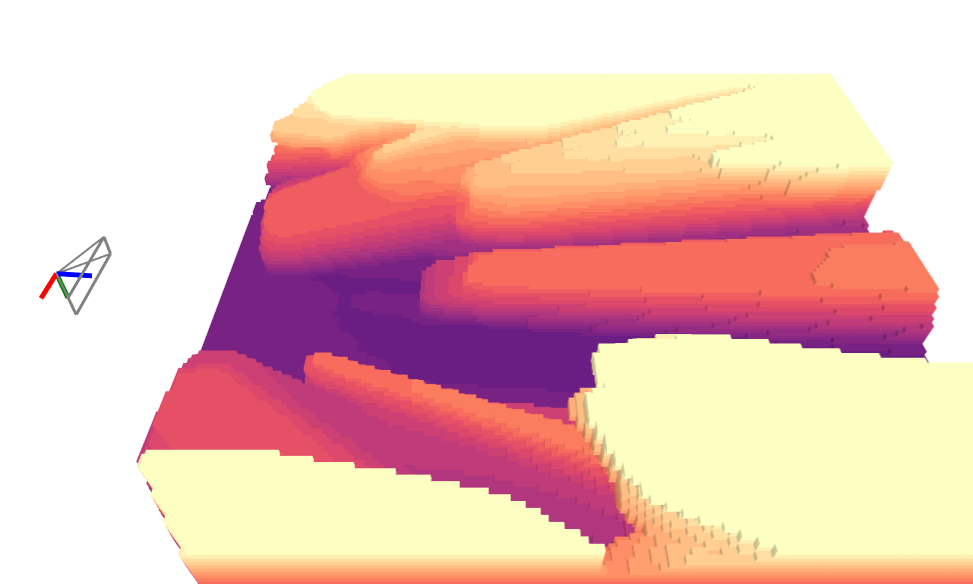} & 
        \includegraphics[width=\linewidth, trim=0cm 0cm 0cm 3cm, clip]{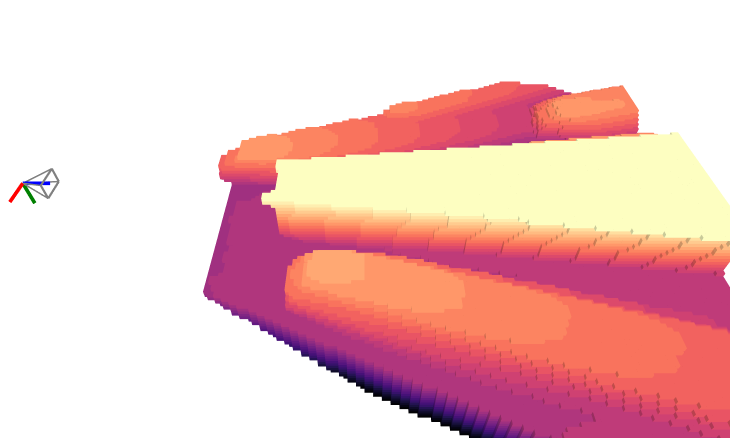} & 
        \includegraphics[width=\linewidth, trim=0cm 0cm 0cm 3cm, clip]{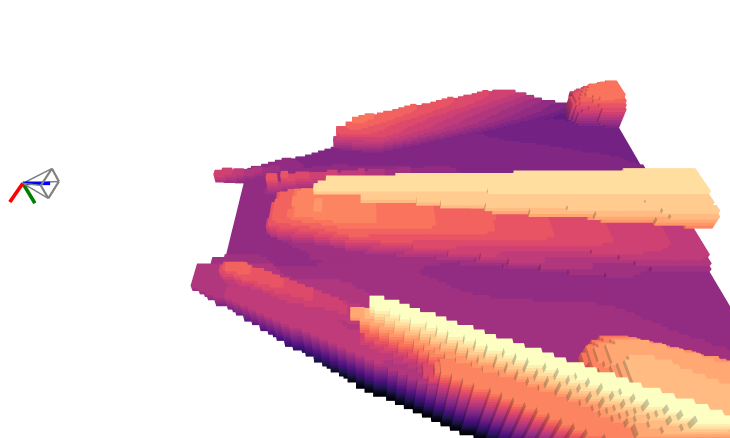} \\

        \rotatebox{90}{\small BTS \cite{wimbauer2023behind}} & 
        \includegraphics[width=\linewidth, trim=0cm 0cm 0cm 3cm, clip]{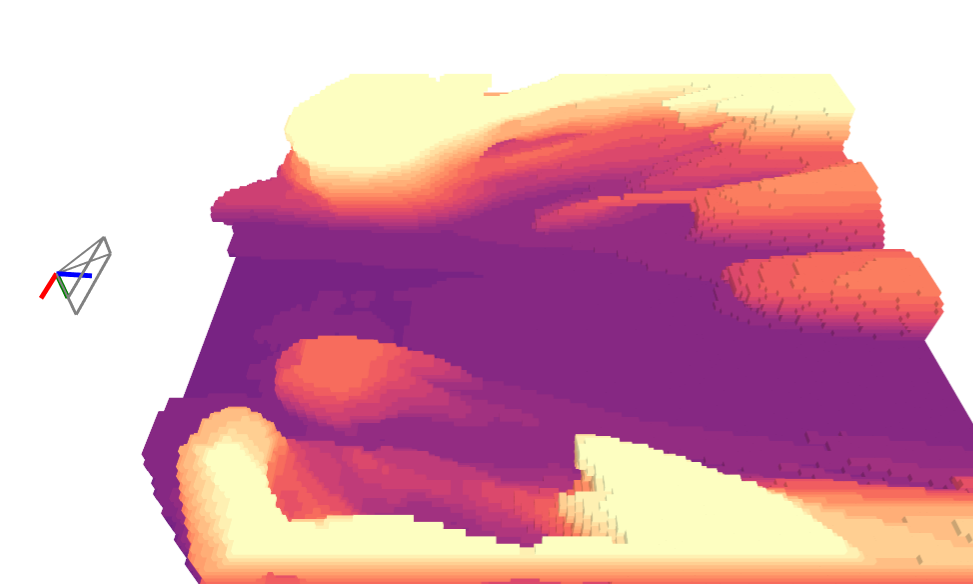} & 
        \includegraphics[width=\linewidth, trim=0cm 0cm 0cm 3cm, clip]{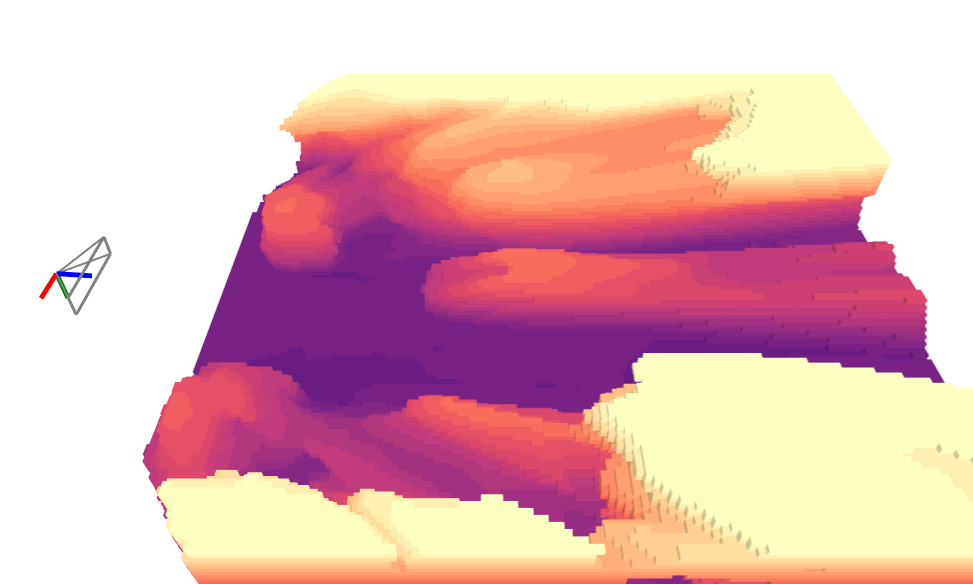} & 
        \includegraphics[width=\linewidth, trim=0cm 0cm 0cm 3cm, clip]{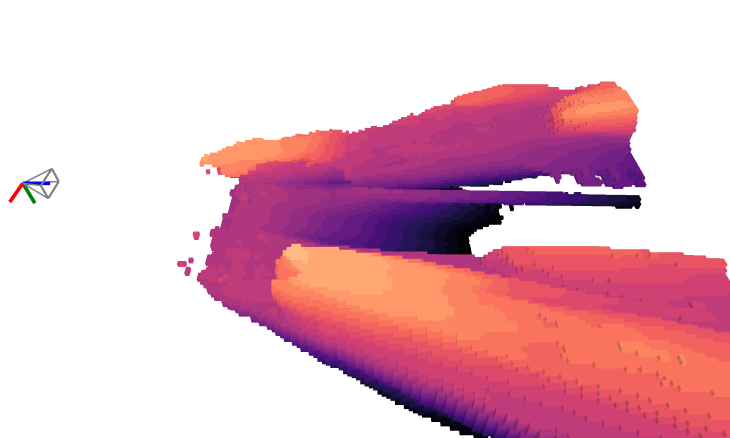} & 
        \includegraphics[width=\linewidth, trim=0cm 0cm 0cm 3cm, clip]{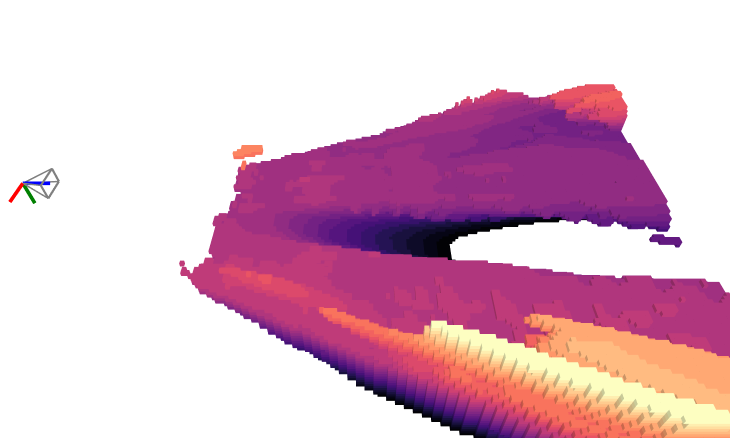} \\

        \rotatebox{90}{\small BTS-D} & 
        \includegraphics[width=\linewidth, trim=0cm 0cm 0cm 3cm, clip]{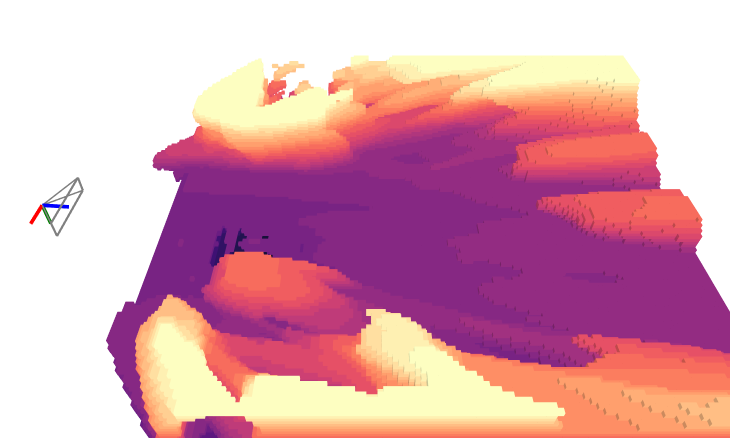} & 
        \includegraphics[width=\linewidth, trim=0cm 0cm 0cm 3cm, clip]{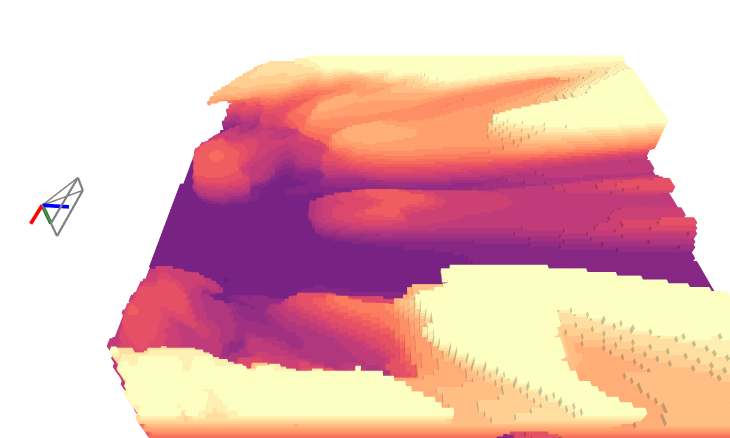} & 
        \includegraphics[width=\linewidth, trim=0cm 0cm 0cm 3cm, clip]{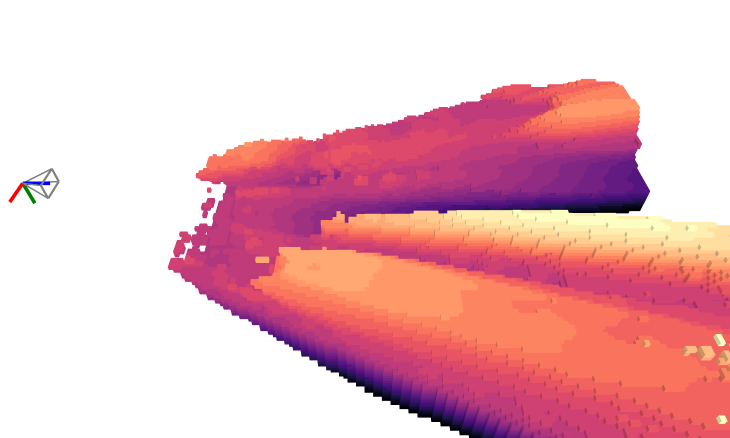} & 
        \includegraphics[width=\linewidth, trim=0cm 0cm 0cm 3cm, clip]{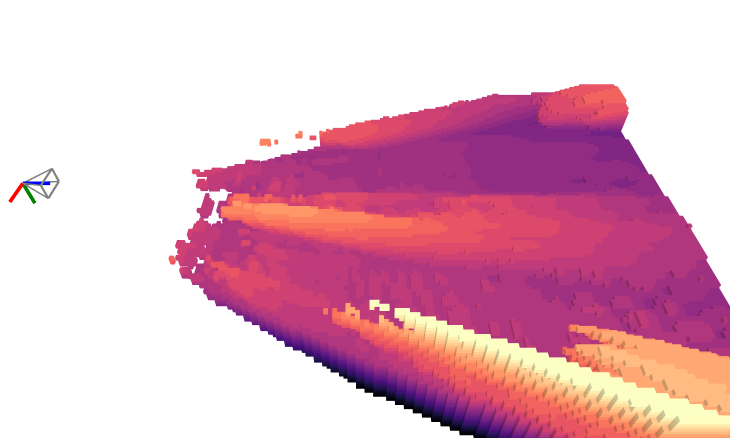} \\

        \rotatebox{90}{\small\textbf{Ours}} & 
        \includegraphics[width=\linewidth, trim=0cm 0cm 0cm 3cm, clip]{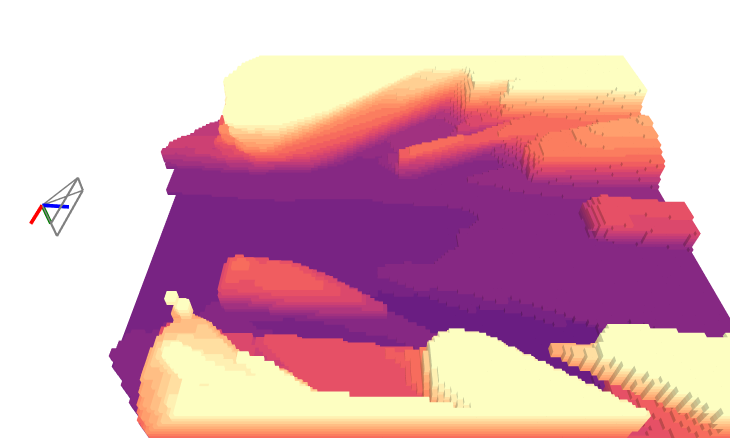} & 
        \includegraphics[width=\linewidth, trim=0cm 0cm 0cm 3cm, clip]{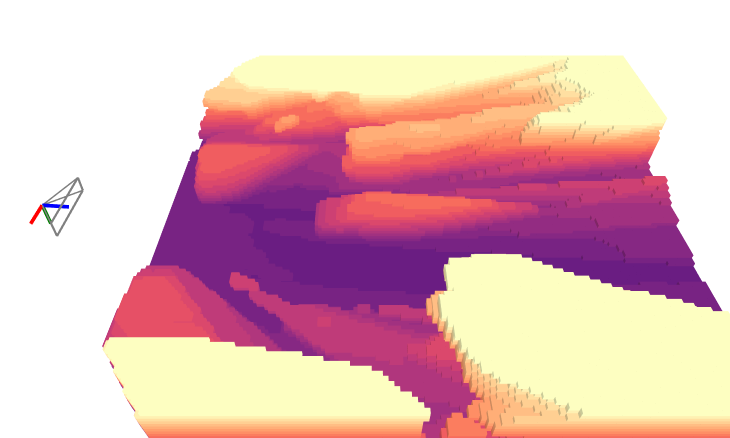} & 
        \includegraphics[width=\linewidth, trim=0cm 0cm 0cm 3cm, clip]{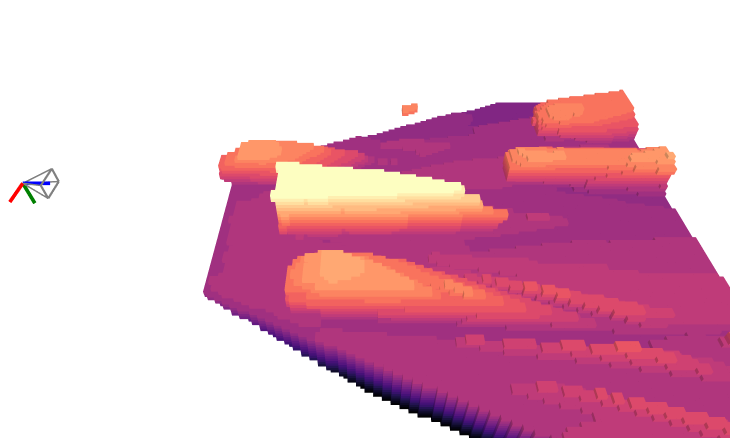} & 
        \includegraphics[width=\linewidth, trim=0cm 0cm 0cm 3cm, clip]{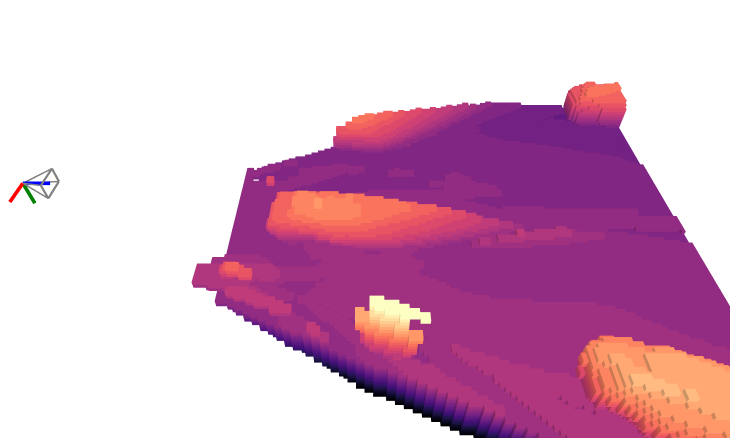} \\

        \rotatebox{90}{\small\textbf{Ours (Distilled)}} & 
        \includegraphics[width=\linewidth, trim=0cm 0cm 0cm 3cm, clip]{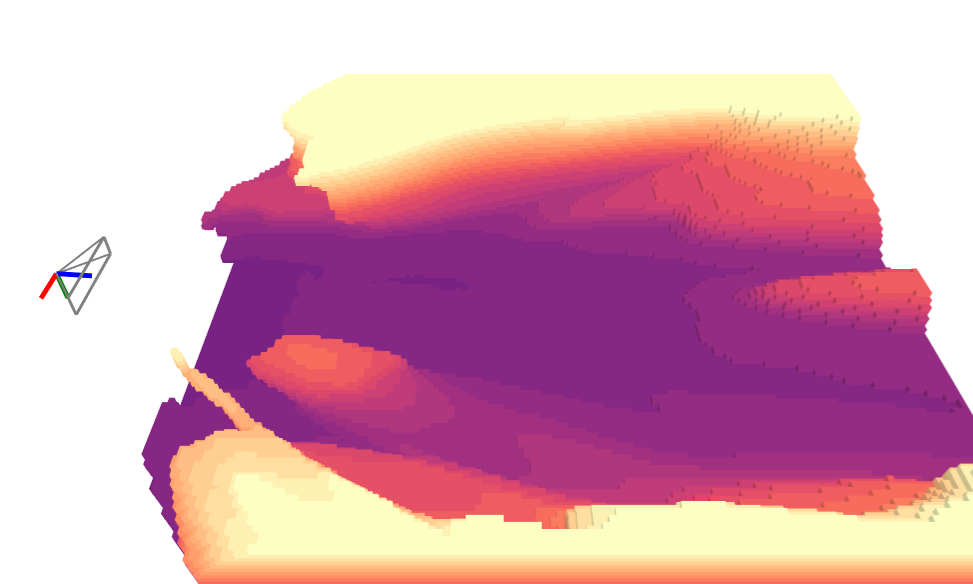} & 
        \includegraphics[width=\linewidth, trim=0cm 0cm 0cm 3cm, clip]{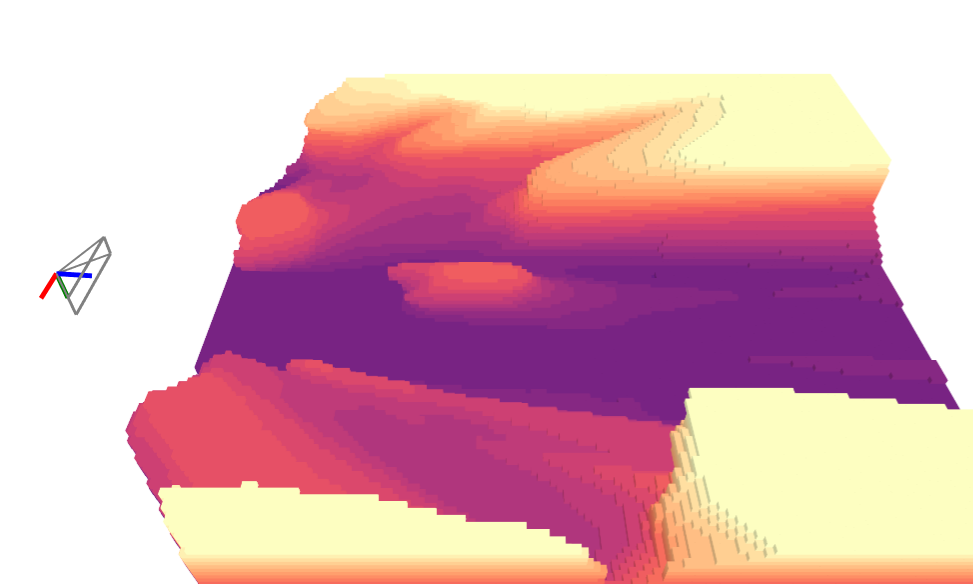} & 
        \includegraphics[width=\linewidth, trim=0cm 0cm 0cm 3cm, clip]{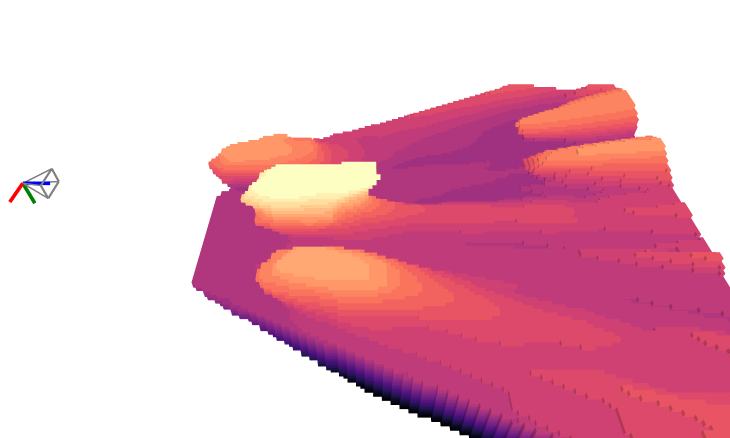} & 
        \includegraphics[width=\linewidth, trim=0cm 0cm 0cm 3cm, clip]{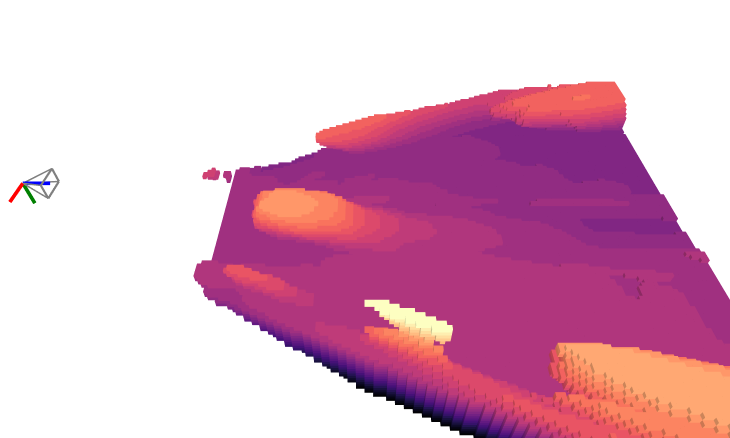} \\
    \end{tabular}
    \caption{\textbf{Qualitative scene reconstruction.} We present scene reconstructions from several models on KITTI-360 \cite{liao2022kitti360} (left) and Waymo \cite{sun2020waymo} (right). The geometry is discretized into a voxel grid with dimensions $x=[-9m,9m],\ y=[-0.25m,4m],\ z=[4m,25m]$. The color in the plots indicates the y-coordinate (up-axis). Despite being trained on single images only, our method achieves accurate reconstructions overall. In addition, it reliably reconstructs dynamic objects, and the distilled model effectively avoids artifacts.
    }
    \label{fig:baselines_qualitative}
    \vspace{-.3cm}
\end{figure*}

We visualize the scene geometry by discretizing the occupancy into a voxel grid.
As illustrated in \cref{fig:baselines_qualitative}, both the synthesized scenes and the geometry predicted by the \recon exhibit high overall quality.
However, the synthesized scenes occasionally display artifacts caused by poor view completion or depth prediction.
The distilled \recon does not exhibit these issues, as it is trained on a diverse dataset.
We contend that, despite being slightly outperformed in quantitative metrics by the directly synthesized geometry, the distilled model is more reliable and significantly faster.
As shown in the 3rd and 4th columns, both BTS variants fail to reconstruct the dynamic object.
This failure stems from their use of multi-view data across multiple timesteps, which introduces inconsistency when the object is in motion.
In contrast, our synthesized scenes naturally handle such scenarios.

\subsection{Scene Synthesis using the \Refmabbr}

% \todo{All interesting stuff for controlnet, depth alignment, etc. should go here.}
% The quality of synthesized novel views has a significant impact on the 3D reconstruction quality of the pseudo-volume \todo{implicit occupancy field (IOF)}. We evaluate the color and depth reconstruction performance of the RGB refinement model (\cref{sec:rgb_image_refinement}) and the impact of our methods for occlusion detection (\cref{sec:occlusion_detection}) and for novel view depth alignment \cref{sec:depth_alignment}. We measure the impact on occupancy reconstruction by ablating our pseudo-volume configuration (\cref{sec:pseudo_volume_results}) and the switching the novel view sampling strategy (\cref{sec:viewpoint_sampling_results}).
The process of synthesizing scene geometry -- by generating novel views and fusing depth maps -- relies on multiple components that significantly influence the final result’s quality.
To this end, we rigorously validate our design choices through a series of ablation studies.

\subsubsection{Conditioning of the \Refmabbr}
\label{sec:rgb_image_refinement}

Our \Refmabbr receives a range of conditioning inputs.
% Warped RGB, the (inverse) depth, and an inpainting mask.
% The inpainting mask is obtained through morphological closing of the occlusion mask and we fill in the RGB with random noise in the occluded areas.
% Furthermore, we feed in the CLIP embedding of the input view.
To assess the impact of conditioning, we compare our full \Refmabbr model against a simplified variant, denoted as $\text{\Refmabbr}^{\text{simple}}$.
This simplified configuration receives only the RGB input, where the region to be inpainted is masked with zeros instead of noise.
We evaluate both configurations on the KITTI-360 dataset ($\text{\Refmabbr}^{\text{K}}$), and further examine the effect of finetuning on Waymo ($\text{\Refmabbr}^{\text{K$\rightarrow$W}}$).
% We compare a \textit{base} and \textit{full} configuration of our RGB image refinement model and train the full model on KITTI-360 (\textit{Full-K}) and finetune it further on Waymo (\textit{Full-W}). The base model receives a $512\times768$ conditioning image with RGB channels and zero-masked areas for inpainting, no CLIP image embedding and is initialized from SD-2.1 weights. In the full model, . We feed a CLIP embedding of the input image to the diffusion model and the ControlNet and initialize the model from SD-2.1-unCLIP. 

As shown in \cref{tab:controlnet_ablations_short}, the full model consistently outperforms the baseline in both color and depth reconstruction metrics on KITTI-360.
On Waymo, finetuning clearly enhances reconstruction quality even further.
\begin{table}[]
    \centering
    \small
\begin{tabular}{lcccc}
    \toprule
    & \multicolumn{2}{c}{Input View} & \multicolumn{2}{c}{Novel View} \\
        \cmidrule(lr){2-3} \cmidrule(lr){4-5}
    \textit{Method} & PSNR ↑ & \makecell{Abs Rel ↓} & PSNR ↑ & \makecell{Abs Rel ↓} \\
    \midrule
    {\footnotesize\textit{KITTI-360}} & & & & \\
    % \small{\textit{KITTI-360}} & & & & \\
    % \multirow{6}{*}{\rotatebox[origin=c]{90}{\small{KITTI-360}}} & Base-old & 18.481 & 0.106 & 20.165 & 0.122 \\
    % & Base-new & 16.514 & 0.24 & -- & -- \\
    $\text{\Refmabbr}^{\text{simple}}$ & 20.197 & 0.174 & 21.323 & 0.154 \\
    % & Full-K-old & 22.527 & 0.093 & 21.803 & 0.110 \\
    % & Full-K-new & 15.585 & 0.272 & -- & -- \\
    \rowcolor{verylightgrey} $\text{\Refmabbr}^{\text{K}}$ & \textbf{23.908} & \textbf{0.161} & \textbf{23.394} & \textbf{0.135} \\
    \midrule
    {\footnotesize\textit{Waymo}} & & & & \\
    % \multirow{2}{*}{\rotatebox[origin=c]{90}{\small{Waymo}}} & Full-K-old & 27.222 & 0.436 & 27.108 & 0.140\\
    % & Full-K-n & 19.748 & 0.322 &  & \\
    $\text{\Refmabbr}^{\text{K}}$ & 26.963 & 0.286 & 21.432 & 0.177 \\
    \rowcolor{verylightgrey} $\text{\Refmabbr}^{\text{K$\rightarrow$W}}$ & \textbf{28.781} & \textbf{0.260} & \textbf{22.603} & \textbf{0.135} \\
    \bottomrule
\end{tabular}

        \vspace{-.3cm}
    \caption{\textbf{Effect of \Refmabbr conditioning.} We evaluate different model configurations and training setups. $\text{\Refmabbr}^{\text{simple}}$ receives only masked RGB input. $\text{\Refmabbr}^{\text{K}}$ and $\text{\Refmabbr}^{\text{K$\rightarrow$W}}$ denote the full model trained on KITTI-360 and finetuned on Waymo, respectively.}
    \label{tab:controlnet_ablations_short}
    \vspace{-.5cm}
\end{table}
% We provide qualitative results to demonstrate the improvement of the full refinement model over its base variant in \cref{fig:controlnet_qualitative_comparison}. Compared to the base model, the full model outputs smoother textures for visible surfaces that resemble those in the input image better and inpainted areas contain fewer artifacts with more coherence between inpainted content and the rest of the image. The full model also avoids the color changes observed within dark regions of the conditioning image when using the base model.
\begin{figure}
    \centering
    \includegraphics[width=\linewidth]{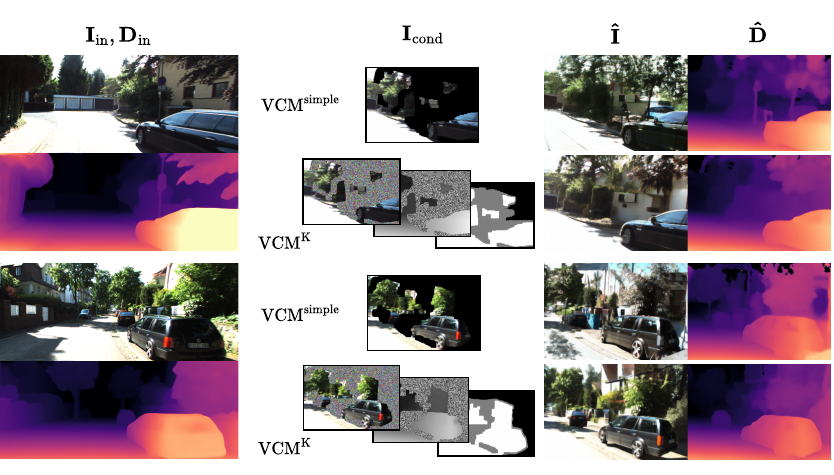}
    \vspace{-.7cm}
    \caption{\textbf{Effect of \Refmabbr conditioning.} See \cref{tab:controlnet_ablations_short}.}
    \label{fig:controlnet_qualitative_comparison}
    \vspace{-.2cm}
\end{figure}

\subsubsection{Occlusion detection in novel views}
\label{sec:occlusion_detection}

\begin{figure}
    \centering
    \includegraphics[width=\linewidth]{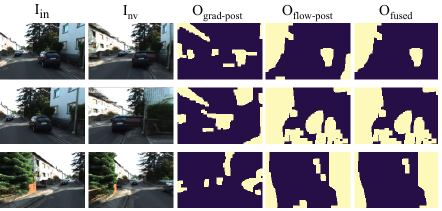}
        % \vspace{-.3cm}
    \caption{\textbf{Effect of occlusion detection strategies.} 
    We compare gradient-based (ours) to optical-flow based occlusion detection, and a combination of both.
    Occlusions are shown in \textcolor{paleYellow}{\colorbox{aubergine}{yellow}}. 
    %at $192\times 640$ resolution. The cropped input patch is used as the reference image for the optical-flow-based occlusion detection. %\todo{remove grad and flow + put I(ref) to the left + remove 'post' from name + space between imgs and masks}
    }
    \label{fig:occlusions_qualitative}
            \vspace{-.3cm}
\end{figure}

The accuracy and robustness of our occlusion detection strategy directly influence the effectiveness of refining incomplete novel views using \Refmabbr.
We compare our depth-gradient-based method against an alternative approach that leverages two-way optical flow between the input and novel views to identify regions visible only in the latter, as well as a conservative fusion of both techniques.
As shown in \cref{fig:occlusions_qualitative}, the depth gradient method robustly captures occlusions without requiring extensive post-filtering.
In contrast, the optical flow approach introduces numerous false positives in the presence of large warps, necessitating aggressive filtering that subsequently reduces true positive detections.
Moreover, it incorrectly classifies all pixels outside the input view’s field of view as occluded.
The fused strategy mitigates some of the false positives compared to optical flow alone but still inherits many of its limitations.

% \begin{table}[]
%     \centering
%     \include{tables/occlusions}
%     \caption{\textbf{Qualitative occlusion detection strategies.} strategies during inference in the novel view. at $192\times 640$ resolution}
%     \label{tab:occlusions_quantitative}
% \end{table}
% We visualize the occlusion maps for different methods before and after filtering in \cref{fig:occlusions_qualitative}. The depth gradient approach captures all occlusions robustly without needing strong post-filtering. In comparison, the optical flow method introduces many false positives for larger warps, necessitating strong filtering which diminishes true positives. It also identifies all pixels outside the field of view of the reference image as occluded. The fused strategy reduces some false positives compared to pure optical flow, but inherits its challenges. 

% \subsubsection{Novel view depth alignment}
% \label{sec:depth_alignment}

% \todo{probably put into appendix}

% Scale alignment of predicted novel view depths is critical for accurate 3D reconstruction. We show that our inverse-median based alignment strategy achieves optimal results for depth reconstruction compared to baselines. Our results in \cref{todo} indicate that alignment via the median depth outperforms other methods. We found that using inverse depth scaling can enhance alignment for objects closer to the camera but introduces larger errors for distant objects. 

% \subsubsection{Pseudo volume configuration}
% \label{sec:pseudo_volume_results}

% \todo{need to rerun some evals right now before writing this... possibly merge this with the viewpoint sampling section}

% The pseudo volume fuses density and color from multiple encoding views. We evaluate this setting by encoding the pseudo volume with both images of KITTI-360's stereo camera and rendering synthetic views. The 3D reconstruction of the scene results from encoding the pseudo volume with both input frames and all synthetic views. As shown in \cref{tab:pseudo_volume}, sampling colors close to surfaces works better than using the mean across all cameras or all cameras that see it. Enforcing the input depth is also better. \todo{rewrite this a little nicer}

% \begin{table}[]
%     \centering
%     \include{tables/pseudo_volume}
%     \caption{configuration of the pseudo volume with stereo inputs on KITTI-360}
%     \label{tab:pseudo_volume}
% \end{table}

\subsubsection{Viewpoint sampling}
\label{sec:viewpoint_sampling_results}

The selection of poses for novel views has a substantial impact on 3D scene geometry reconstruction, especially in occluded regions.
We evaluate several pose sampling strategies using our occupancy reconstruction benchmark.
As shown in \cref{tab:pose_sampling_quantitative}, translating the camera along the input baseline by $[-3m, 3m]$ yields better results than orbiting it by $[-10^\circ, 10^\circ]$ around a vertical axis at a depth of $5m$. 
Camera rigs outperform random warp sampling strategies, as their cameras are already defined in sensible poses. 
% \todo{The 3D consistency among novel views and between novel and input views is crucial when reconstructing from synthetic multi-view data. TODO possibly add info on enforcing the input depth here if we merge in the previous chapter}
% \todo{qualitative voxel grid plot with to samplers}
\begin{table}[]
    \centering
    \small
\begin{tabular}{lcccc}
\toprule
\textit{Pose sampling} & $||\mathbb{P}||$ & \makecell{$\text{O}_\text{acc}$ ↑} & \makecell{$\text{IE}_\text{acc}$ ↑} & \makecell{$\text{IE}_\text{rec}$ ↑}\\
\midrule
Random orbit & 4 & 0.923 & 0.674 & 0.596 \\
% Random orbit 8 keep 6 & 0.924 & 0.684 & 0.622 \\
Random shift & 4 & 0.927 & 0.692 & 0.636 \\
% Random shift 8 keep 6 & 0.930 & 0.704 & 0.677 \\
Rig & 4 & 0.930 & 0.708 & 0.687 \\
% Rig-8 keep 4 & 0.931 & 0.714 & 0.705 \\
\rowcolor{verylightgrey} \textbf{Rig} & \textbf{8} & \textbf{0.932} & \textbf{0.720 }& \textbf{0.729} \\

\bottomrule
\end{tabular}

        \vspace{-.3cm}
    \caption{\textbf{Pose sampling strategies.} We evaluate the quality of generated geometry under various pose sampling strategies. A camera rig with eight predefined poses and random rotations achieves the best performance.}
    \label{tab:pose_sampling_quantitative}
    \vspace{-.5cm}
\end{table}
% \begin{figure}
%     \centering
%     \setlength{\tabcolsep}{2pt} % Adjust the space between columns
%     \begin{tabular}{
%     >{\centering\arraybackslash}m{.49\linewidth} 
%     >{\centering\arraybackslash}m{.49\linewidth}}
%         \multicolumn{2}{c}{\includegraphics[width=.5\linewidth]{example-image}} \\
%         \includegraphics[width=\linewidth]{example-image} & 
%         \includegraphics[width=\linewidth]{example-image} \\
%         rig6of8 & 
%         worse strat \\
%     \end{tabular}
%     \caption{Visualization of sampling methods. \todo{add vis for sampling strat showing novel views for the views and then the pseudo vol}}
%     \label{fig:enter-label}
% \end{figure}

\subsection{Distillation into a Feed-Forward Model}
% \todo{All stuff related to the reconstructor should go here.}

\begin{figure}
    \centering
    \small
    \setlength{\tabcolsep}{0.5pt} % Adjust the space between columns
    \begin{tabular}{m{.05\linewidth} >{\centering\arraybackslash}m{.3\linewidth} >{\centering\arraybackslash}m{.3\linewidth} >{\centering\arraybackslash}m{.3\linewidth}}
        \rotatebox{90}{Input} & 
        \includegraphics[width=\linewidth]{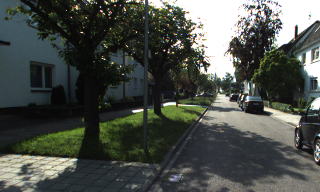} & 
        \includegraphics[width=\linewidth]{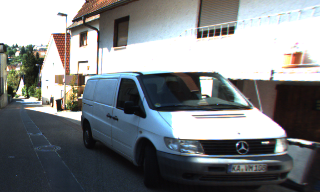} & 
        \includegraphics[width=\linewidth]{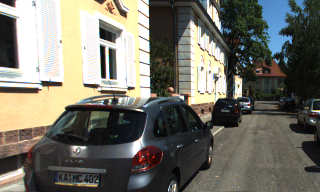} \\

        \rotatebox{90}{\textbf{Full}} & 
        \includegraphics[width=\linewidth]{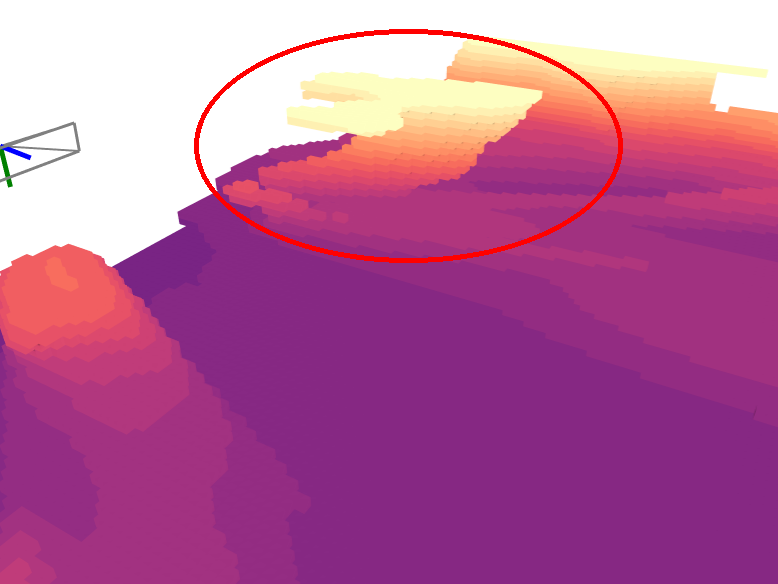} & 
        \includegraphics[width=\linewidth]{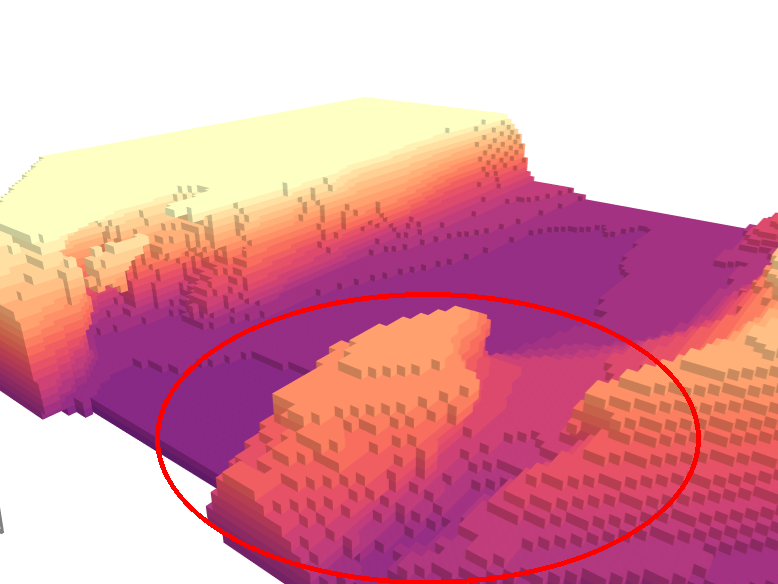} & 
        \includegraphics[width=\linewidth]{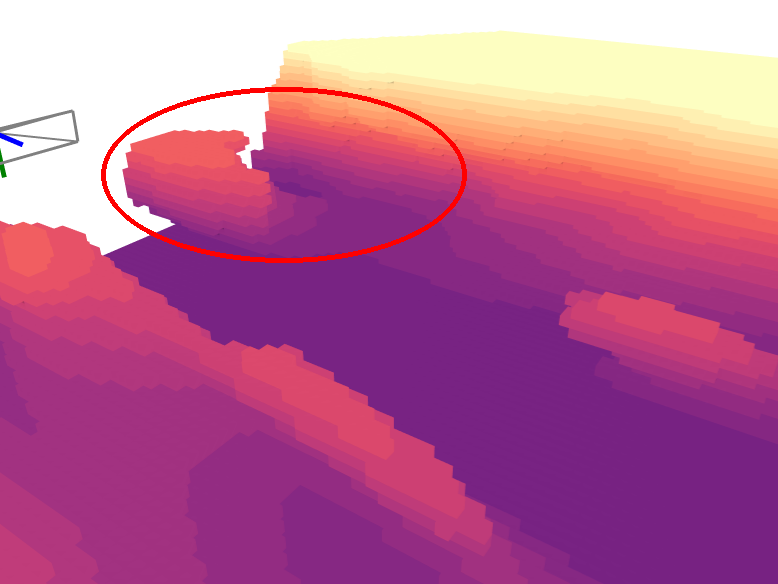} \\

        \rotatebox{90}{w/o $\mathcal{L}_\text{depth}$} & 
        \includegraphics[width=\linewidth]{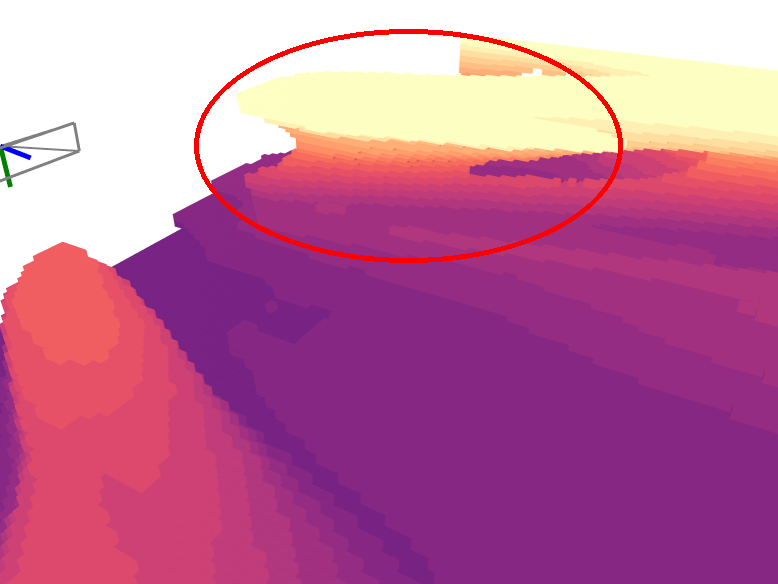} & 
        \includegraphics[width=\linewidth]{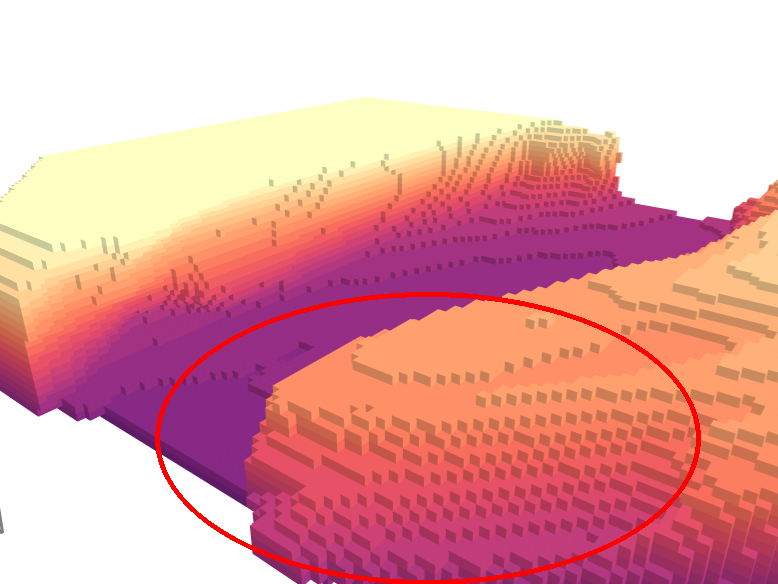} & 
        \includegraphics[width=\linewidth]{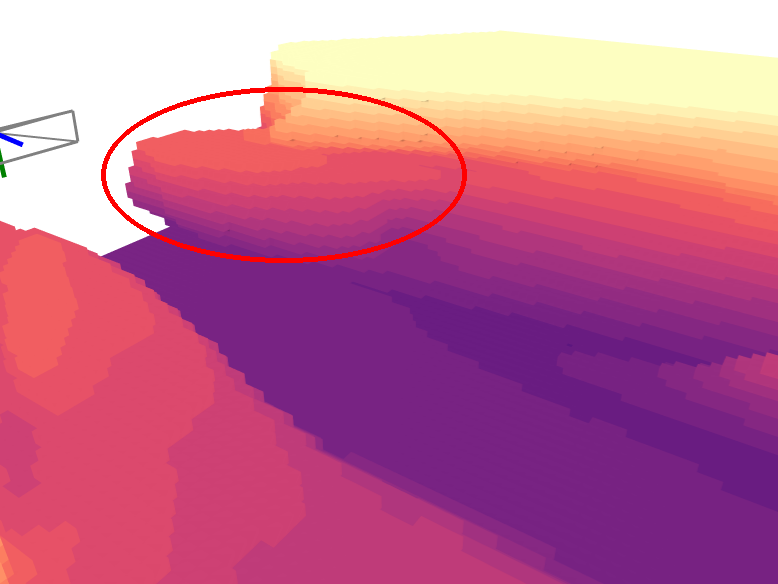} \\
        
        \rotatebox{90}{w/o $\mathcal{L}_\text{occ}$} & 
        \includegraphics[width=\linewidth]{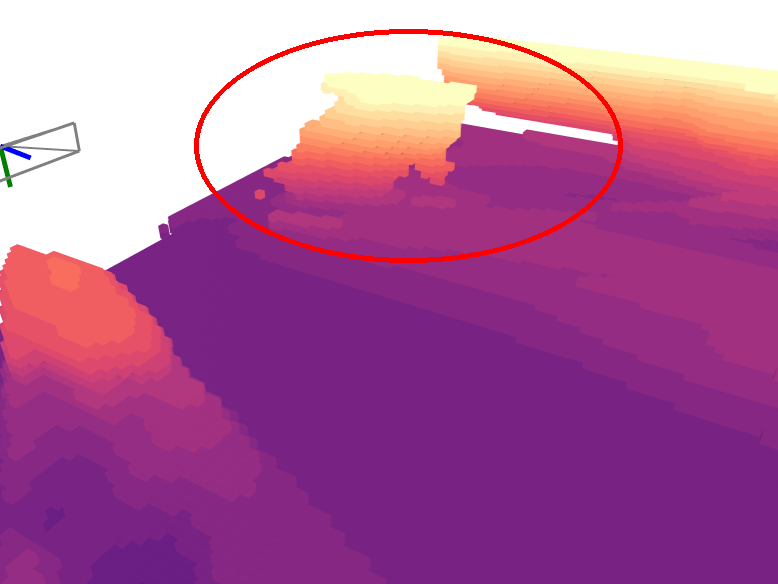} & 
        \includegraphics[width=\linewidth]{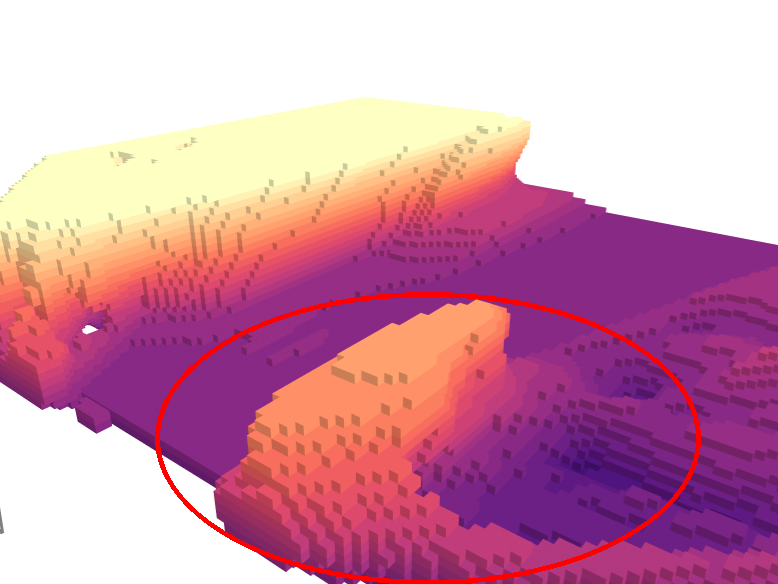} & 
        \includegraphics[width=\linewidth]{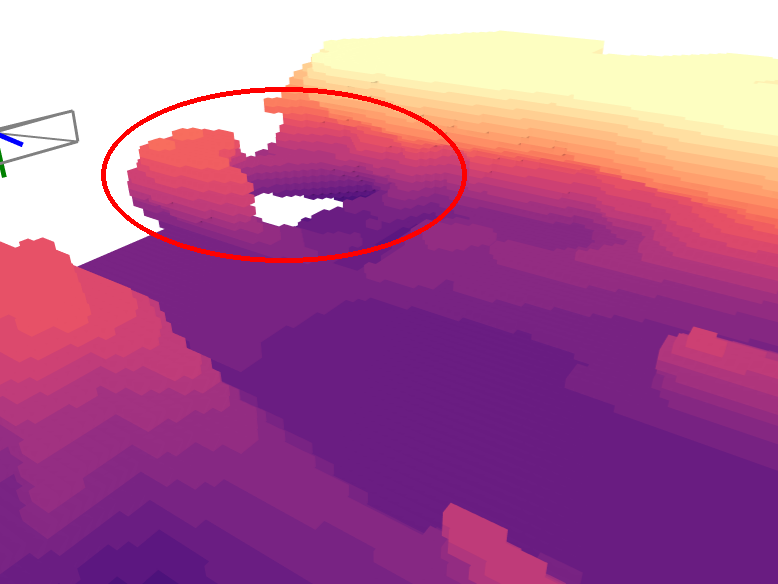} \\

        \rotatebox{90}{w/o uncert.} & 
        \includegraphics[width=\linewidth]{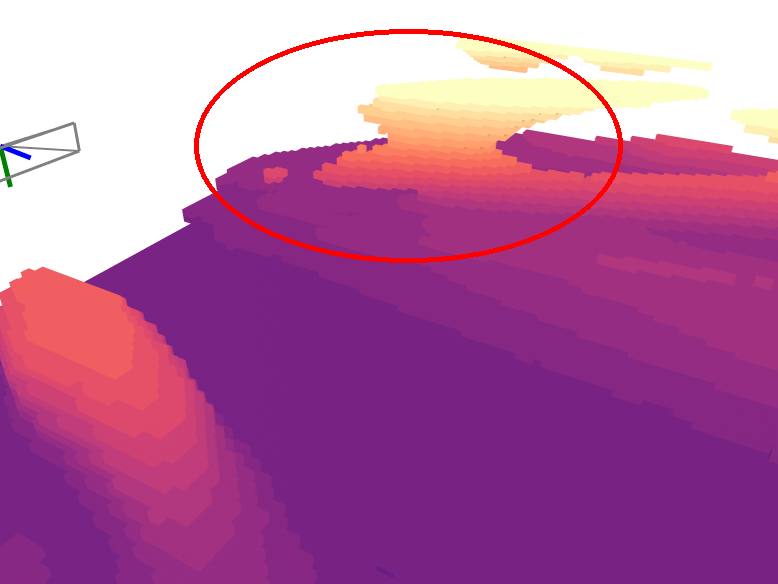} & 
        \includegraphics[width=\linewidth]{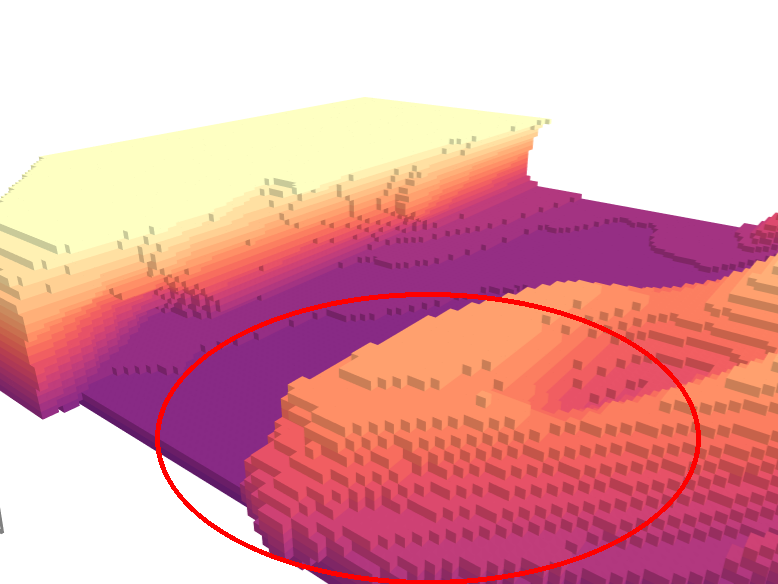} & 
        \includegraphics[width=\linewidth]{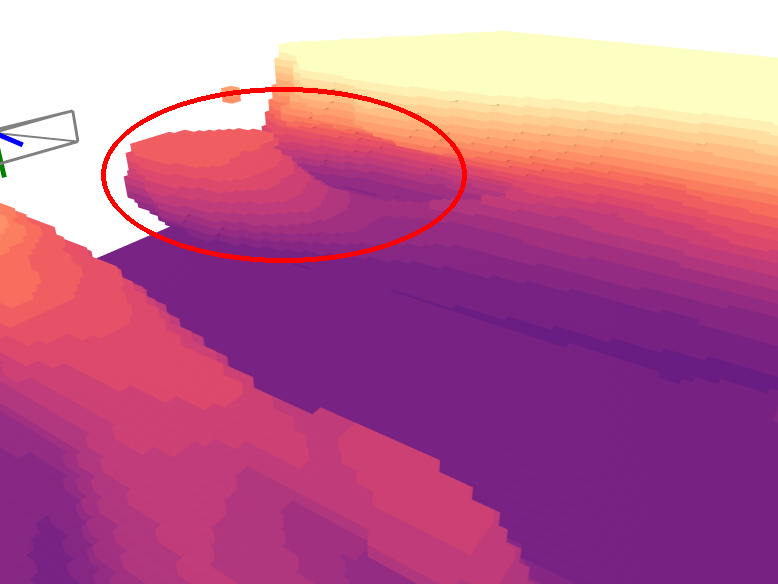} \\
    \end{tabular}
    \vspace{-.3cm}
    \caption{\textbf{Qualitative effect of different loss terms.} See \cref{tab:ablation_study}.}
    \label{fig:ablation_qualitative}
            \vspace{-.3cm}
\end{figure}

The loss configuration used to distill synthetic data into the \recon is critical to its performance. 
% To this end, we report results for ablations of this loss (\cref{sec:loss_ablations}) and the predicted uncertainty of the reconstruction (\cref{sec:uncertainty}).
We assess the contribution of individual loss terms by evaluating occupancy reconstruction metrics in \cref{tab:ablation_study} and providing qualitative comparisons in \cref{fig:ablation_qualitative}.
% We mask out all rays in $\mathcal{L}_\text{depth}$ where 80\% of the sampled density is invalid as well as rays where the GT depth is outside of the learned values. 
% For $\mathcal{L}_\text{depth-nv}$ we also discard all rays that do not pass through occlusions. 
% Compared to the ablations, our \textit{full} loss achieves competitive $\text{O}_\text{acc}$ and the best $\text{IE}_\text{acc}$. The invisible and empty reconstruction recall $\text{IE}_\text{rec}$ increases when removing $\mathcal{L}_\text{occ}$, but $\text{IE}_\text{acc}$ goes down.
% This suggests that the model simply predicts more free space, even though this might often not be correct.
Compared to ablated variants, our \textit{full} loss setup achieves competitive $\text{O}_\text{acc}$ and the highest $\text{IE}_\text{acc}$.
While removing $\mathcal{L}_\text{occ}$ increases the invisible and empty region recall ($\text{IE}_\text{rec}$), it lowers $\text{IE}_\text{acc}$ -- indicating that the model over-predicts free space, which may not be correct.
% \cref{fig:ablation_qualitative} shows that removing $\mathcal{L}_\text{depth}$ results in less accurate predictions of invisible and empty regions. Without $\mathcal{L}_\text{occ}$, the reconstruction of invisible, but non-empty space is worse, which is not captured by $\text{IE}_\text{rec}$. 
Moreover, the predicted uncertainty is needed to balance the depth and volumetric loss w.r.t.\ each other.
As illustrated in \cref{fig:ablation_qualitative}, omitting $\mathcal{L}_\text{depth}$ leads to less accurate reconstruction of invisible and empty regions.
Eliminating $\mathcal{L}_\text{occ}$ worsens the reconstruction of invisible but non-empty space, an effect not reflected in $\text{IE}_\text{rec}$.
Finally, the predicted uncertainty is essential for balancing the depth and volumetric losses relative to one another.
\begin{table}[]
    \centering
    \small
\begin{tabular}{lllll}
    \toprule
     Method & $\text{O}_\text{acc}$ ↑ & $\text{IE}_\text{acc}$ ↑ & $\text{IE}_\text{rec}$ ↑\\
    \midrule
    \rowcolor{verylightgrey} \textbf{Full} & \underline{0.902} & \textbf{0.705} & \underline{0.706} \\
    w/o $\mathcal{L}_\text{depth}$ & 0.857 & 0.656 & 0.443 \\
    w/o $\mathcal{L}_\text{occ}$ & \textbf{0.907} & \underline{0.699} & \textbf{0.757} \\
    w/o uncertainty & \textbf{0.907} & 0.697 & 0.701 \\
    \bottomrule
\end{tabular}

        \vspace{-.3cm}
    \caption{\textbf{Quantitative effect of different loss terms.} We show reconstructions produced by models with different loss terms turned off. Overall, the full model produces the most convincing results.}
            \vspace{-.3cm}
    \label{tab:ablation_study}
\end{table}

% \paragraph{Reconstruction uncertainty} The predicted uncertainty does not only balance the loss function, but is also interpretable. As shown in the birds-eye-view visualizations of 3D reconstructions and uncertainty in \cref{fig:uncertainty_profiles}, the uncertainty is low in all visible regions and high in invisible ones. For large solid objects, such as houses, the uncertainty decreases some distance behind the depth.

% \begin{figure}
%     \centering
%     \setlength{\tabcolsep}{2pt} % Adjust the space between columns
%     \begin{tabular}{m{.49\linewidth} >{\centering\arraybackslash}m{.49\linewidth}}
%         \includegraphics[width=\linewidth]{figures/uncertainty_profiles/input_054.png} & 
%         \includegraphics[width=\linewidth]{figures/uncertainty_profiles/profile_054_full.png} \\
%         \includegraphics[width=\linewidth]{figures/uncertainty_profiles/input_119.png} & 
%         \includegraphics[width=\linewidth]{figures/uncertainty_profiles/profile_119_full.png} \\
%     \end{tabular}
%     \caption{Uncertainty profiles}
%     \label{fig:uncertainty_profiles}
% \end{figure}
\section{Conclusion}
\label{sec:conclusion}

In this work, we proposed a novel approach to monocular 3D reconstruction that can be trained using only single images.
The resulting geometry matches or exceeds the performance of models trained with multi-view supervision.

% Furthermore, a novel pose sampling strategy was developed to optimize for views that enhance the understanding of scene structure, subsequently improving the quality of 3D reconstructions.
% \paragraph{Limitations} The approach presented faces several limitations. The process of synthesizing novel view ground truths is computationally expensive, such that we faced scalability issues when training the monocular reconstructor. Our cascaded reconstruction technique, while effective, requires significant memory resources, restricting its application in resource-constrained environments. Additionally, the quality of synthetic novel views may not consistently reach the fidelity required to use pixel value-based reconstruction losses effectively, necessitating further investigation. Finally, the results we reached for the distillation of the synthetic novel views into the reconstructor leave opportunity for improvement.

% \paragraph{Future work}

% - gaussian splatting instead to pseudo vol to get synthetic 3D GT

% - RGBD diffusion model -> joint RGBD refinement

% - try to learn color as well and focus on novel view synthesis

% - try to integrate this framework to improve scene level 3d recon with nerf or gaussian splatting

{\small
\noindent\textbf{Acknowledgements:} This work was funded by the ERC Advanced Grant ”SIMULACRON” (agreement \#884679), the GNI Project ”AI4Twinning”, and the DFG project CR 250/26-1 ”4D YouTube”.
}

{
    \small
    \bibliographystyle{ieeenat_fullname}
    \bibliography{main}

\begin{thebibliography}{68}
\providecommand{\natexlab}[1]{#1}
\providecommand{\url}[1]{\texttt{#1}}
\expandafter\ifx\csname urlstyle\endcsname\relax
  \providecommand{\doi}[1]{doi: #1}\else
  \providecommand{\doi}{doi: \begingroup \urlstyle{rm}\Url}\fi

\bibitem[Bhat et~al.(2023)Bhat, Birkl, Wofk, Wonka, and M{\"u}ller]{bhat2023zoedepth}
Shariq~Farooq Bhat, Reiner Birkl, Diana Wofk, Peter Wonka, and Matthias M{\"u}ller.
\newblock Zoedepth: Zero-shot transfer by combining relative and metric depth.
\newblock \emph{arXiv preprint arXiv:2302.12288}, 2023.

\bibitem[Bochkovskii et~al.(2024)Bochkovskii, Delaunoy, Germain, Santos, Zhou, Richter, and Koltun]{bochkovskii2024depth}
Aleksei Bochkovskii, Ama{\"e}l Delaunoy, Hugo Germain, Marcel Santos, Yichao Zhou, Stephan~R Richter, and Vladlen Koltun.
\newblock Depth pro: Sharp monocular metric depth in less than a second.
\newblock \emph{arXiv preprint arXiv:2410.02073}, 2024.

\bibitem[Cai et~al.(2023)Cai, Chan, Peng, Shahbazi, Obukhov, Van~Gool, and Wetzstein]{cai2022diffdreamer}
Shengqu Cai, Eric~Ryan Chan, Songyou Peng, Mohamad Shahbazi, Anton Obukhov, Luc Van~Gool, and Gordon Wetzstein.
\newblock Diffdreamer: Towards consistent unsupervised single-view scene extrapolation with conditional diffusion models.
\newblock In \emph{ICCV}, 2023.

\bibitem[Cao and De~Charette(2022)]{cao2022monoscene}
Anh-Quan Cao and Raoul De~Charette.
\newblock Monoscene: Monocular 3d semantic scene completion.
\newblock In \emph{Proceedings of the IEEE/CVF Conference on Computer Vision and Pattern Recognition}, pages 3991--4001, 2022.

\bibitem[Cao and De~Charette(2023)]{cao2023scenerf}
Anh-Quan Cao and Raoul De~Charette.
\newblock Scenerf: Self-supervised monocular 3d scene reconstruction with radiance fields.
\newblock In \emph{Proceedings of the IEEE/CVF International Conference on Computer Vision}, pages 9387--9398, 2023.

\bibitem[Chen et~al.(2023)Chen, Yu, Ge, Yao, Xie, Wu, Wang, Kwok, Luo, Lu, et~al.]{chen2023pixart}
Junsong Chen, Jincheng Yu, Chongjian Ge, Lewei Yao, Enze Xie, Yue Wu, Zhongdao Wang, James Kwok, Ping Luo, Huchuan Lu, et~al.
\newblock Pixart-$\alpha$: Fast training of diffusion transformer for photorealistic text-to-image synthesis.
\newblock \emph{arXiv preprint arXiv:2310.00426}, 2023.

\bibitem[Chung et~al.(2023)Chung, Lee, Nam, Lee, and Lee]{chung2023luciddreamer}
Jaeyoung Chung, Suyoung Lee, Hyeongjin Nam, Jaerin Lee, and Kyoung~Mu Lee.
\newblock Luciddreamer: Domain-free generation of 3d gaussian splatting scenes.
\newblock \emph{arXiv preprint arXiv:2311.13384}, 2023.

\bibitem[Eigen and Fergus(2015)]{eigen2015predicting}
David Eigen and Rob Fergus.
\newblock Predicting depth, surface normals and semantic labels with a common multi-scale convolutional architecture.
\newblock In \emph{Proceedings of the IEEE international conference on computer vision}, pages 2650--2658, 2015.

\bibitem[Eigen et~al.(2014)Eigen, Puhrsch, and Fergus]{eigen2014depth}
David Eigen, Christian Puhrsch, and Rob Fergus.
\newblock Depth map prediction from a single image using a multi-scale deep network.
\newblock \emph{Advances in neural information processing systems}, 27, 2014.

\bibitem[Engstler et~al.(2024)Engstler, Vedaldi, Laina, and Rupprecht]{engstler2024invisible}
Paul Engstler, Andrea Vedaldi, Iro Laina, and Christian Rupprecht.
\newblock Invisible stitch: Generating smooth 3d scenes with depth inpainting.
\newblock \emph{arXiv preprint arXiv:2404.19758}, 2024.

\bibitem[Fu et~al.(2018)Fu, Gong, Wang, Batmanghelich, and Tao]{fu2018deep}
Huan Fu, Mingming Gong, Chaohui Wang, Kayhan Batmanghelich, and Dacheng Tao.
\newblock Deep ordinal regression network for monocular depth estimation.
\newblock In \emph{Proceedings of the IEEE conference on computer vision and pattern recognition}, pages 2002--2011, 2018.

\bibitem[Gao et~al.(2024)Gao, Holynski, Henzler, Brussee, Martin-Brualla, Srinivasan, Barron, and Poole]{gao2024cat3d}
Ruiqi Gao, Aleksander Holynski, Philipp Henzler, Arthur Brussee, Ricardo Martin-Brualla, Pratul Srinivasan, Jonathan~T Barron, and Ben Poole.
\newblock Cat3d: Create anything in 3d with multi-view diffusion models.
\newblock \emph{arXiv preprint arXiv:2405.10314}, 2024.

\bibitem[Godard et~al.(2017)Godard, Mac~Aodha, and Brostow]{godard2017unsupervised}
Cl{\'e}ment Godard, Oisin Mac~Aodha, and Gabriel~J Brostow.
\newblock Unsupervised monocular depth estimation with left-right consistency.
\newblock In \emph{Proceedings of the IEEE conference on computer vision and pattern recognition}, pages 270--279, 2017.

\bibitem[Godard et~al.(2019)Godard, Mac~Aodha, Firman, and Brostow]{godard2019digging}
Cl{\'e}ment Godard, Oisin Mac~Aodha, Michael Firman, and Gabriel~J Brostow.
\newblock Digging into self-supervised monocular depth estimation.
\newblock In \emph{Proceedings of the IEEE/CVF international conference on computer vision}, pages 3828--3838, 2019.

\bibitem[Guizilini et~al.(2021)Guizilini, Ambrus, Burgard, and Gaidon]{guizilini2021sparse}
Vitor Guizilini, Rares Ambrus, Wolfram Burgard, and Adrien Gaidon.
\newblock Sparse auxiliary networks for unified monocular depth prediction and completion.
\newblock In \emph{Proceedings of the ieee/cvf conference on computer vision and pattern recognition}, pages 11078--11088, 2021.

\bibitem[Hayler et~al.(2024)Hayler, Wimbauer, Muhle, Rupprecht, and Cremers]{hayler2024s4c}
Adrian Hayler, Felix Wimbauer, Dominik Muhle, Christian Rupprecht, and Daniel Cremers.
\newblock S4c: Self-supervised semantic scene completion with neural fields.
\newblock In \emph{2024 International Conference on 3D Vision (3DV)}, pages 409--420. IEEE, 2024.

\bibitem[H{\"o}llein et~al.(2023)H{\"o}llein, Cao, Owens, Johnson, and Nie{\ss}ner]{hollein2023text2room}
Lukas H{\"o}llein, Ang Cao, Andrew Owens, Justin Johnson, and Matthias Nie{\ss}ner.
\newblock Text2room: Extracting textured 3d meshes from 2d text-to-image models.
\newblock In \emph{Proceedings of the IEEE/CVF International Conference on Computer Vision}, pages 7909--7920, 2023.

\bibitem[H{\"o}llein et~al.(2024)H{\"o}llein, Bo{\v{z}}i{\v{c}}, M{\"u}ller, Novotny, Tseng, Richardt, Zollh{\"o}fer, and Nie{\ss}ner]{hollein2024viewdiff}
Lukas H{\"o}llein, Alja{\v{z}} Bo{\v{z}}i{\v{c}}, Norman M{\"u}ller, David Novotny, Hung-Yu Tseng, Christian Richardt, Michael Zollh{\"o}fer, and Matthias Nie{\ss}ner.
\newblock Viewdiff: 3d-consistent image generation with text-to-image models.
\newblock In \emph{Proceedings of the IEEE/CVF conference on computer vision and pattern recognition}, pages 5043--5052, 2024.

\bibitem[Hong et~al.(2023)Hong, Zhang, Gu, Bi, Zhou, Liu, Liu, Sunkavalli, Bui, and Tan]{hong2023lrm}
Yicong Hong, Kai Zhang, Jiuxiang Gu, Sai Bi, Yang Zhou, Difan Liu, Feng Liu, Kalyan Sunkavalli, Trung Bui, and Hao Tan.
\newblock Lrm: Large reconstruction model for single image to 3d.
\newblock \emph{arXiv preprint arXiv:2311.04400}, 2023.

\bibitem[Hu et~al.(2024)Hu, Yin, Zhang, Cai, Long, Chen, Wang, Yu, Shen, and Shen]{hu2024metric3d}
Mu Hu, Wei Yin, Chi Zhang, Zhipeng Cai, Xiaoxiao Long, Hao Chen, Kaixuan Wang, Gang Yu, Chunhua Shen, and Shaojie Shen.
\newblock Metric3d v2: A versatile monocular geometric foundation model for zero-shot metric depth and surface normal estimation.
\newblock \emph{IEEE Transactions on Pattern Analysis and Machine Intelligence}, 2024.

\bibitem[Huang et~al.(2023)Huang, Zheng, Zhang, Zhou, and Lu]{huang2023tri}
Yuanhui Huang, Wenzhao Zheng, Yunpeng Zhang, Jie Zhou, and Jiwen Lu.
\newblock Tri-perspective view for vision-based 3d semantic occupancy prediction.
\newblock In \emph{Proceedings of the IEEE/CVF conference on computer vision and pattern recognition}, pages 9223--9232, 2023.

\bibitem[Huang et~al.(2024)Huang, Zheng, Zhang, Zhou, and Lu]{huang2024selfocc}
Yuanhui Huang, Wenzhao Zheng, Borui Zhang, Jie Zhou, and Jiwen Lu.
\newblock Selfocc: Self-supervised vision-based 3d occupancy prediction.
\newblock In \emph{Proceedings of the IEEE/CVF Conference on Computer Vision and Pattern Recognition}, pages 19946--19956, 2024.

\bibitem[Kajiya and Von~Herzen(1984)]{kajiya1984ray}
James~T Kajiya and Brian~P Von~Herzen.
\newblock Ray tracing volume densities.
\newblock \emph{ACM SIGGRAPH computer graphics}, 18\penalty0 (3):\penalty0 165--174, 1984.

\bibitem[Karnewar et~al.(2023)Karnewar, Vedaldi, Novotny, and Mitra]{karnewar2023holodiffusion}
Animesh Karnewar, Andrea Vedaldi, David Novotny, and Niloy~J Mitra.
\newblock Holodiffusion: Training a 3d diffusion model using 2d images.
\newblock In \emph{Proceedings of the IEEE/CVF conference on computer vision and pattern recognition}, pages 18423--18433, 2023.

\bibitem[Kendall and Gal(2017)]{kendall2017uncertainties}
Alex Kendall and Yarin Gal.
\newblock What uncertainties do we need in bayesian deep learning for computer vision?
\newblock \emph{Advances in neural information processing systems}, 30, 2017.

\bibitem[Laina et~al.(2016)Laina, Rupprecht, Belagiannis, Tombari, and Navab]{laina2016deeper}
Iro Laina, Christian Rupprecht, Vasileios Belagiannis, Federico Tombari, and Nassir Navab.
\newblock Deeper depth prediction with fully convolutional residual networks.
\newblock In \emph{2016 Fourth international conference on 3D vision (3DV)}, pages 239--248. IEEE, 2016.

\bibitem[Li et~al.(2024)Li, Fischer, Segu, Pollefeys, Van~Gool, and Tombari]{li2024know}
Rui Li, Tobias Fischer, Mattia Segu, Marc Pollefeys, Luc Van~Gool, and Federico Tombari.
\newblock Know your neighbors: Improving single-view reconstruction via spatial vision-language reasoning.
\newblock In \emph{Proceedings of the IEEE/CVF Conference on Computer Vision and Pattern Recognition}, pages 9848--9858, 2024.

\bibitem[Li et~al.(2023{\natexlab{a}})Li, Yu, Choy, Xiao, Alvarez, Fidler, Feng, and Anandkumar]{li2023voxformer}
Yiming Li, Zhiding Yu, Christopher Choy, Chaowei Xiao, Jose~M Alvarez, Sanja Fidler, Chen Feng, and Anima Anandkumar.
\newblock Voxformer: Sparse voxel transformer for camera-based 3d semantic scene completion.
\newblock In \emph{Proceedings of the IEEE/CVF conference on computer vision and pattern recognition}, pages 9087--9098, 2023{\natexlab{a}}.

\bibitem[Li et~al.(2022)Li, Wang, Snavely, and Kanazawa]{li2022infinitenature}
Zhengqi Li, Qianqian Wang, Noah Snavely, and Angjoo Kanazawa.
\newblock Infinitenature-zero: Learning perpetual view generation of natural scenes from single images.
\newblock In \emph{European Conference on Computer Vision}, pages 515--534. Springer, 2022.

\bibitem[Li et~al.(2023{\natexlab{b}})Li, Yu, Austin, Fang, Lan, Kautz, and Alvarez]{li2023fb}
Zhiqi Li, Zhiding Yu, David Austin, Mingsheng Fang, Shiyi Lan, Jan Kautz, and Jose~M Alvarez.
\newblock Fb-occ: 3d occupancy prediction based on forward-backward view transformation.
\newblock \emph{arXiv preprint arXiv:2307.01492}, 2023{\natexlab{b}}.

\bibitem[Liao et~al.(2022)Liao, Xie, and Geiger]{liao2022kitti360}
Yiyi Liao, Jun Xie, and Andreas Geiger.
\newblock {KITTI}-360: A novel dataset and benchmarks for urban scene understanding in 2d and 3d.
\newblock \emph{Pattern Analysis and Machine Intelligence (PAMI)}, 2022.

\bibitem[Liu et~al.(2021)Liu, Tucker, Jampani, Makadia, Snavely, and Kanazawa]{liu2021infinite}
Andrew Liu, Richard Tucker, Varun Jampani, Ameesh Makadia, Noah Snavely, and Angjoo Kanazawa.
\newblock Infinite nature: Perpetual view generation of natural scenes from a single image.
\newblock In \emph{Proceedings of the IEEE/CVF International Conference on Computer Vision}, pages 14458--14467, 2021.

\bibitem[Liu et~al.(2015)Liu, Shen, Lin, and Reid]{liu2015learning}
Fayao Liu, Chunhua Shen, Guosheng Lin, and Ian Reid.
\newblock Learning depth from single monocular images using deep convolutional neural fields.
\newblock \emph{IEEE transactions on pattern analysis and machine intelligence}, 38\penalty0 (10):\penalty0 2024--2039, 2015.

\bibitem[Liu et~al.(2023)Liu, Wu, Van~Hoorick, Tokmakov, Zakharov, and Vondrick]{liu2023zero}
Ruoshi Liu, Rundi Wu, Basile Van~Hoorick, Pavel Tokmakov, Sergey Zakharov, and Carl Vondrick.
\newblock Zero-1-to-3: Zero-shot one image to 3d object.
\newblock In \emph{Proceedings of the IEEE/CVF international conference on computer vision}, pages 9298--9309, 2023.

\bibitem[Miao et~al.(2023)Miao, Liu, Chen, Gong, Xu, Hu, and Zhou]{miao2023occdepth}
Ruihang Miao, Weizhou Liu, Mingrui Chen, Zheng Gong, Weixin Xu, Chen Hu, and Shuchang Zhou.
\newblock Occdepth: A depth-aware method for 3d semantic scene completion.
\newblock \emph{arXiv preprint arXiv:2302.13540}, 2023.

\bibitem[Mildenhall et~al.(2021)Mildenhall, Srinivasan, Tancik, Barron, Ramamoorthi, and Ng]{mildenhall2021nerf}
Ben Mildenhall, Pratul~P Srinivasan, Matthew Tancik, Jonathan~T Barron, Ravi Ramamoorthi, and Ren Ng.
\newblock Nerf: Representing scenes as neural radiance fields for view synthesis.
\newblock \emph{Communications of the ACM}, 65\penalty0 (1):\penalty0 99--106, 2021.

\bibitem[Pan et~al.(2024)Pan, Liu, Zhang, Huang, Li, Xie, Wang, Liu, and Zhang]{pan2024renderocc}
Mingjie Pan, Jiaming Liu, Renrui Zhang, Peixiang Huang, Xiaoqi Li, Hongwei Xie, Bing Wang, Li Liu, and Shanghang Zhang.
\newblock Renderocc: Vision-centric 3d occupancy prediction with 2d rendering supervision.
\newblock In \emph{2024 IEEE International Conference on Robotics and Automation (ICRA)}, pages 12404--12411. IEEE, 2024.

\bibitem[Piccinelli et~al.(2024)Piccinelli, Yang, Sakaridis, Segu, Li, Van~Gool, and Yu]{piccinelli2024unidepth}
Luigi Piccinelli, Yung-Hsu Yang, Christos Sakaridis, Mattia Segu, Siyuan Li, Luc Van~Gool, and Fisher Yu.
\newblock Unidepth: Universal monocular metric depth estimation.
\newblock In \emph{Proceedings of the IEEE/CVF Conference on Computer Vision and Pattern Recognition}, pages 10106--10116, 2024.

\bibitem[Podell et~al.(2023)Podell, English, Lacey, Blattmann, Dockhorn, M{\"u}ller, Penna, and Rombach]{podell2023sdxl}
Dustin Podell, Zion English, Kyle Lacey, Andreas Blattmann, Tim Dockhorn, Jonas M{\"u}ller, Joe Penna, and Robin Rombach.
\newblock Sdxl: Improving latent diffusion models for high-resolution image synthesis.
\newblock \emph{arXiv preprint arXiv:2307.01952}, 2023.

\bibitem[Poole et~al.(2022)Poole, Jain, Barron, and Mildenhall]{poole2022dreamfusion}
Ben Poole, Ajay Jain, Jonathan~T Barron, and Ben Mildenhall.
\newblock Dreamfusion: Text-to-3d using 2d diffusion.
\newblock \emph{arXiv preprint arXiv:2209.14988}, 2022.

\bibitem[Qian et~al.(2023)Qian, Mai, Hamdi, Ren, Siarohin, Li, Lee, Skorokhodov, Wonka, Tulyakov, et~al.]{qian2023magic123}
Guocheng Qian, Jinjie Mai, Abdullah Hamdi, Jian Ren, Aliaksandr Siarohin, Bing Li, Hsin-Ying Lee, Ivan Skorokhodov, Peter Wonka, Sergey Tulyakov, et~al.
\newblock Magic123: One image to high-quality 3d object generation using both 2d and 3d diffusion priors.
\newblock \emph{arXiv preprint arXiv:2306.17843}, 2023.

\bibitem[Radford et~al.(2021)Radford, Kim, Hallacy, Ramesh, Goh, Agarwal, Sastry, Askell, Mishkin, Clark, et~al.]{radford2021clip}
Alec Radford, Jong~Wook Kim, Chris Hallacy, Aditya Ramesh, Gabriel Goh, Sandhini Agarwal, Girish Sastry, Amanda Askell, Pamela Mishkin, Jack Clark, et~al.
\newblock Learning transferable visual models from natural language supervision.
\newblock In \emph{International conference on machine learning}, pages 8748--8763. PmLR, 2021.

\bibitem[Ramesh et~al.(2021)Ramesh, Pavlov, Goh, Gray, Voss, Radford, Chen, and Sutskever]{ramesh2021zero}
Aditya Ramesh, Mikhail Pavlov, Gabriel Goh, Scott Gray, Chelsea Voss, Alec Radford, Mark Chen, and Ilya Sutskever.
\newblock Zero-shot text-to-image generation.
\newblock In \emph{International conference on machine learning}, pages 8821--8831. Pmlr, 2021.

\bibitem[Ranftl et~al.(2020)Ranftl, Lasinger, Hafner, Schindler, and Koltun]{ranftl2020towards}
Ren{\'e} Ranftl, Katrin Lasinger, David Hafner, Konrad Schindler, and Vladlen Koltun.
\newblock Towards robust monocular depth estimation: Mixing datasets for zero-shot cross-dataset transfer.
\newblock \emph{IEEE transactions on pattern analysis and machine intelligence}, 44\penalty0 (3):\penalty0 1623--1637, 2020.

\bibitem[Ranftl et~al.(2021)Ranftl, Bochkovskiy, and Koltun]{ranftl2021vision}
Ren{\'e} Ranftl, Alexey Bochkovskiy, and Vladlen Koltun.
\newblock Vision transformers for dense prediction.
\newblock In \emph{Proceedings of the IEEE/CVF international conference on computer vision}, pages 12179--12188, 2021.

\bibitem[Roessle et~al.(2022)Roessle, Barron, Mildenhall, Srinivasan, and Nie{\ss}ner]{roessle2022dense}
Barbara Roessle, Jonathan~T Barron, Ben Mildenhall, Pratul~P Srinivasan, and Matthias Nie{\ss}ner.
\newblock Dense depth priors for neural radiance fields from sparse input views.
\newblock In \emph{Proceedings of the IEEE/CVF Conference on Computer Vision and Pattern Recognition}, pages 12892--12901, 2022.

\bibitem[Rombach et~al.(2022)Rombach, Blattmann, Lorenz, Esser, and Ommer]{rombach2022high}
Robin Rombach, Andreas Blattmann, Dominik Lorenz, Patrick Esser, and Bj{\"o}rn Ommer.
\newblock High-resolution image synthesis with latent diffusion models.
\newblock In \emph{Proceedings of the IEEE/CVF conference on computer vision and pattern recognition}, pages 10684--10695, 2022.

\bibitem[Saharia et~al.(2022)Saharia, Chan, Saxena, Li, Whang, Denton, Ghasemipour, Gontijo~Lopes, Karagol~Ayan, Salimans, et~al.]{saharia2022photorealistic}
Chitwan Saharia, William Chan, Saurabh Saxena, Lala Li, Jay Whang, Emily~L Denton, Kamyar Ghasemipour, Raphael Gontijo~Lopes, Burcu Karagol~Ayan, Tim Salimans, et~al.
\newblock Photorealistic text-to-image diffusion models with deep language understanding.
\newblock \emph{Advances in neural information processing systems}, 35:\penalty0 36479--36494, 2022.

\bibitem[Schult et~al.(2024)Schult, Tsai, H{\"o}llein, Wu, Wang, Ma, Li, Wang, Wimbauer, He, et~al.]{schult2024controlroom3d}
Jonas Schult, Sam Tsai, Lukas H{\"o}llein, Bichen Wu, Jialiang Wang, Chih-Yao Ma, Kunpeng Li, Xiaofang Wang, Felix Wimbauer, Zijian He, et~al.
\newblock Controlroom3d: Room generation using semantic proxy rooms.
\newblock In \emph{Proceedings of the IEEE/CVF Conference on Computer Vision and Pattern Recognition}, pages 6201--6210, 2024.

\bibitem[Shi et~al.(2023)Shi, Wang, Ye, Long, Li, and Yang]{shi2023mvdream}
Yichun Shi, Peng Wang, Jianglong Ye, Mai Long, Kejie Li, and Xiao Yang.
\newblock Mvdream: Multi-view diffusion for 3d generation.
\newblock \emph{arXiv preprint arXiv:2308.16512}, 2023.

\bibitem[Shriram et~al.(2024)Shriram, Trevithick, Liu, and Ramamoorthi]{shriram2024realmdreamer}
Jaidev Shriram, Alex Trevithick, Lingjie Liu, and Ravi Ramamoorthi.
\newblock Realmdreamer: Text-driven 3d scene generation with inpainting and depth diffusion.
\newblock \emph{arXiv preprint arXiv:2404.07199}, 2024.

\bibitem[Siddiqui et~al.(2025)Siddiqui, Monnier, Kokkinos, Kariya, Kleiman, Garreau, Gafni, Neverova, Vedaldi, Shapovalov, et~al.]{siddiqui2025meta}
Yawar Siddiqui, Tom Monnier, Filippos Kokkinos, Mahendra Kariya, Yanir Kleiman, Emilien Garreau, Oran Gafni, Natalia Neverova, Andrea Vedaldi, Roman Shapovalov, et~al.
\newblock Meta 3d assetgen: Text-to-mesh generation with high-quality geometry, texture, and pbr materials.
\newblock \emph{Advances in Neural Information Processing Systems}, 37:\penalty0 9532--9564, 2025.

\bibitem[Sima et~al.(2023)Sima, Tong, Wang, Chen, Wu, Deng, Gu, Lu, Luo, Lin, et~al.]{sima2023scene}
Chonghao Sima, Wenwen Tong, Tai Wang, Li Chen, Silei Wu, Hanming Deng, Yi Gu, Lewei Lu, Ping Luo, Dahua Lin, et~al.
\newblock Scene as occupancy.
\newblock \emph{arXiv preprint arXiv:2306.02851}, 2023.

\bibitem[Sun et~al.(2020)Sun, Kretzschmar, Dotiwalla, Chouard, Patnaik, Tsui, Guo, Zhou, Chai, Caine, Vasudevan, Han, Ngiam, Zhao, Timofeev, Ettinger, Krivokon, Gao, Joshi, Zhang, Shlens, Chen, and Anguelov]{sun2020waymo}
Pei Sun, Henrik Kretzschmar, Xerxes Dotiwalla, Aurelien Chouard, Vijaysai Patnaik, Paul Tsui, James Guo, Yin Zhou, Yuning Chai, Benjamin Caine, Vijay Vasudevan, Wei Han, Jiquan Ngiam, Hang Zhao, Aleksei Timofeev, Scott Ettinger, Maxim Krivokon, Amy Gao, Aditya Joshi, Yu Zhang, Jonathon Shlens, Zhifeng Chen, and Dragomir Anguelov.
\newblock Scalability in perception for autonomous driving: Waymo open dataset.
\newblock In \emph{Proceedings of the IEEE/CVF Conference on Computer Vision and Pattern Recognition (CVPR)}, 2020.

\bibitem[Szymanowicz et~al.(2023)Szymanowicz, Rupprecht, and Vedaldi]{szymanowicz2023viewset}
Stanislaw Szymanowicz, Christian Rupprecht, and Andrea Vedaldi.
\newblock Viewset diffusion:(0-) image-conditioned 3d generative models from 2d data.
\newblock In \emph{Proceedings of the IEEE/CVF international conference on computer vision}, pages 8863--8873, 2023.

\bibitem[Tang et~al.(2024)Tang, Chen, Chen, Wang, Zeng, and Liu]{tang2024lgm}
Jiaxiang Tang, Zhaoxi Chen, Xiaokang Chen, Tengfei Wang, Gang Zeng, and Ziwei Liu.
\newblock Lgm: Large multi-view gaussian model for high-resolution 3d content creation.
\newblock In \emph{European Conference on Computer Vision}, pages 1--18. Springer, 2024.

\bibitem[Wang et~al.(2015)Wang, Shen, Lin, Cohen, Price, and Yuille]{wang2015towards}
Peng Wang, Xiaohui Shen, Zhe Lin, Scott Cohen, Brian Price, and Alan~L Yuille.
\newblock Towards unified depth and semantic prediction from a single image.
\newblock In \emph{Proceedings of the IEEE conference on computer vision and pattern recognition}, pages 2800--2809, 2015.

\bibitem[Wang et~al.(2023)Wang, Lu, Wang, Bao, Li, Su, and Zhu]{wang2023prolificdreamer}
Zhengyi Wang, Cheng Lu, Yikai Wang, Fan Bao, Chongxuan Li, Hang Su, and Jun Zhu.
\newblock Prolificdreamer: High-fidelity and diverse text-to-3d generation with variational score distillation.
\newblock \emph{Advances in Neural Information Processing Systems}, 36:\penalty0 8406--8441, 2023.

\bibitem[Wei et~al.(2024)Wei, Zhang, Bi, Tan, Luan, Deschaintre, Sunkavalli, Su, and Xu]{wei2024meshlrm}
Xinyue Wei, Kai Zhang, Sai Bi, Hao Tan, Fujun Luan, Valentin Deschaintre, Kalyan Sunkavalli, Hao Su, and Zexiang Xu.
\newblock Meshlrm: Large reconstruction model for high-quality meshes.
\newblock \emph{arXiv preprint arXiv:2404.12385}, 2024.

\bibitem[Wimbauer et~al.(2023)Wimbauer, Yang, Rupprecht, and Cremers]{wimbauer2023behind}
Felix Wimbauer, Nan Yang, Christian Rupprecht, and Daniel Cremers.
\newblock Behind the scenes: Density fields for single view reconstruction.
\newblock In \emph{Proceedings of the IEEE/CVF Conference on Computer Vision and Pattern Recognition}, pages 9076--9086, 2023.

\bibitem[Xu et~al.(2024)Xu, Shi, Yifan, Chen, Yang, Peng, Shen, and Wetzstein]{xu2024grm}
Yinghao Xu, Zifan Shi, Wang Yifan, Hansheng Chen, Ceyuan Yang, Sida Peng, Yujun Shen, and Gordon Wetzstein.
\newblock Grm: Large gaussian reconstruction model for efficient 3d reconstruction and generation.
\newblock In \emph{European Conference on Computer Vision}, pages 1--20. Springer, 2024.

\bibitem[Yang et~al.(2024)Yang, Kang, Huang, Xu, Feng, and Zhao]{yang2024depth}
Lihe Yang, Bingyi Kang, Zilong Huang, Xiaogang Xu, Jiashi Feng, and Hengshuang Zhao.
\newblock Depth anything: Unleashing the power of large-scale unlabeled data.
\newblock In \emph{Proceedings of the IEEE/CVF Conference on Computer Vision and Pattern Recognition}, pages 10371--10381, 2024.

\bibitem[Yang et~al.(2025)Yang, Kang, Huang, Zhao, Xu, Feng, and Zhao]{yang2025depth}
Lihe Yang, Bingyi Kang, Zilong Huang, Zhen Zhao, Xiaogang Xu, Jiashi Feng, and Hengshuang Zhao.
\newblock Depth anything v2.
\newblock \emph{Advances in Neural Information Processing Systems}, 37:\penalty0 21875--21911, 2025.

\bibitem[Yin et~al.(2021)Yin, Zhang, Wang, Niklaus, Mai, Chen, and Shen]{Wei2021CVPR}
Wei Yin, Jianming Zhang, Oliver Wang, Simon Niklaus, Long Mai, Simon Chen, and Chunhua Shen.
\newblock Learning to recover 3d scene shape from a single image.
\newblock In \emph{Proc. IEEE Conf. Comp. Vis. Patt. Recogn. (CVPR)}, 2021.

\bibitem[Yin et~al.(2023)Yin, Zhang, Chen, Cai, Yu, Wang, Chen, and Shen]{yin2023metric3d}
Wei Yin, Chi Zhang, Hao Chen, Zhipeng Cai, Gang Yu, Kaixuan Wang, Xiaozhi Chen, and Chunhua Shen.
\newblock Metric3d: Towards zero-shot metric 3d prediction from a single image.
\newblock In \emph{Proceedings of the IEEE/CVF International Conference on Computer Vision}, pages 9043--9053, 2023.

\bibitem[Zhang et~al.(2024)Zhang, Li, Wan, Wang, and Liao]{zhang2024text2nerf}
Jingbo Zhang, Xiaoyu Li, Ziyu Wan, Can Wang, and Jing Liao.
\newblock Text2nerf: Text-driven 3d scene generation with neural radiance fields.
\newblock \emph{IEEE Transactions on Visualization and Computer Graphics}, 30\penalty0 (12):\penalty0 7749--7762, 2024.

\bibitem[Zhang et~al.(2023)Zhang, Rao, and Agrawala]{zhang2023controlnet}
Lvmin Zhang, Anyi Rao, and Maneesh Agrawala.
\newblock Adding conditional control to text-to-image diffusion models.
\newblock In \emph{IEEE International Conference on Computer Vision (ICCV)}, 2023.

\bibitem[Zhou et~al.(2024)Zhou, Fan, Xu, Chang, Chari, Bharadwaj, You, Wang, and Kadambi]{zhou2024dreamscene360}
Shijie Zhou, Zhiwen Fan, Dejia Xu, Haoran Chang, Pradyumna Chari, Tejas Bharadwaj, Suya You, Zhangyang Wang, and Achuta Kadambi.
\newblock Dreamscene360: Unconstrained text-to-3d scene generation with panoramic gaussian splatting.
\newblock In \emph{European Conference on Computer Vision}, pages 324--342. Springer, 2024.

\end{thebibliography}
}

\end{document}

\maketitle
\appendix

\noindent
In this supplementary material, we present additional implementation details, extended results, and ablation studies.

\section{Video Results}

We include video results showcasing multiple sequences from both KITTI-360 and Waymo, available on our \href{https://philippwulff.github.io/dream-to-recon}{project website}.
The videos illustrate per-frame volumetric reconstructions for our method alongside several baseline approaches.

\section{Implementation Details}

\paragraph{Model training.}
Our \refm is built upon Stable-Diffusion-2.1-unCLIP, employing the standard ControlNet architecture for conditioning.
For training data synthesis, we render novel views at a resolution of $192\times 360$, followed by re-rendering the input views at $192\times 288$, which are then upscaled to $512\times 768$ to approximate the resolution used during the original training of Stable-Diffusion.
The input image is derived from the re-rendered view.
We use DDIM sampling during training and evaluate the model using UniPCMultiStep sampling with five steps.
The training is initialized from the official Stable-Diffusion weights and runs for 20 epochs with a batch size of 20 and a constant learning rate of $10^{-5}$.
During ControlNet training, UniDepth is used for depth prediction. However, subsequent experiments showed that Metric3D yields more accurate predictions than UniDepth; therefore, we adopt Metric3D for training the \recon model.
To construct the pseudo-volume, we employ 48 coarse samples, 16 fine samples, and 16 ray samples placed within $\sigma=2$ of the estimated depth.

The \recon architecture utilizes the ResNet50-based U-Net backbone from Behind the Scenes.
To ensure stability in mixed-precision training, batch normalization layers are added to the decoder portion of the backbone.
The backbone outputs a frustum-aligned density grid of dimensions $192\times 640\times c$, where $c \in {64, 128}$, with an inverse growth pattern along the $z$-axis.
The resulting occupancy field has a resolution of $Z=64$, inversely spaced between depths of $3m$ and $50m$.

Training is conducted using the UniPCMultiStepScheduler, with automatic mixed precision (AMP), caching, and batch size increases beginning from the second epoch.

\paragraph{Ground truth for evaluation.}
For each timestamp, a 3D occupancy grid is generated by removing all voxels that contain points from 300 consecutive LiDAR sweeps.
The overall reconstruction accuracy ($\text{O}_\text{acc}$) is computed within a cuboid defined by $x = [-4m, 4m]$, $y = [-0.75m, 0m]$, and $z = [4m, 20m]$ in the input camera’s coordinate frame.
Given the higher frequency of dynamic objects in the Waymo dataset, we accumulate 300 LiDAR sweeps and retain occupancy only within the ground truth 3D bounding boxes.
For Waymo, evaluation is performed on the validation set, as 3D bounding box annotations are not available in the test set.

Additionally, the metrics $\text{IE}\text{acc}$ and $\text{IE}\text{rec}$ quantify the reconstruction’s accuracy and recall, respectively.
These metrics are particularly important as they measure reconstruction quality \textit{beyond} what a monocular depth estimation (MDE) model can typically achieve.

\begin{figure*}[t]
    \centering
    \includegraphics[width=\linewidth,trim={0 0 .2cm 0}, clip]{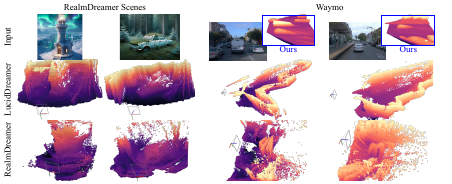}
    \caption{\textbf{Render-refine-repeat Comparison.} Occupancy reconstruction for different RRR methods against ours (marked \textcolor{blue}{blue}).}
    \label{fig:sup_rrr}
\end{figure*}

\section{Comparison With Render-Refine-Repeat Methods}
Depth-based warping and inpainting are also used in other render-refine-repeat (RRR) works.
These approaches generally tackle the Text-to-3D task and focus exclusively on novel view synthesis.
We compare against two SOTA methods \texttt{LucidDreamer} and \texttt{RealmDreamer} (concurrent work), which use a comparable StableDiffusion model as us, and extract the geometry of the scene.
For fairness, we test images from their papers and from Waymo, as seen in \cref{fig:sup_rrr}.
There are to two main differences:
\textbf{1)} Due to the focus on NVS, the resulting geometry is subpar, even for example images from their papers.
Here, our \textit{synthetic occupancy field} comes into play.
\textbf{2)} Because of the focus on Text-to-3D rather than Image-to-3D, existing RRR methods work well for images generated using the diffusion model itself (often in cartoon or fantasy-style).
Here, the diffusion model can inpaint effectively even without finetuning.
However, when using real images, the inpainting produces poor results.
Our novel \textit{VCM} with the corresponding warp-based training solves this issue.

\section{Additional Experiments}

\paragraph{Quantitative effect of occlusions.}
\begin{table}[]
    \centering
    % \footnotesize
    \begin{tabular}{ccccc}
    \toprule
    \makecell{Depth\\gradients} & \makecell{Optical\\flow} & \makecell{PSNR ↑} & \makecell{Abs Rel ↓} & \makecell{RMSE ↓}\\
    \midrule
    \rowcolor{verylightgrey} \cmark & \xmark & \textbf{20.920} & \textbf{0.133} & \textbf{0.093}\\
    \xmark & \cmark & 18.707 & \underline{0.179} & 0.121\\
    \cmark & \cmark & \underline{18.933} & \underline{0.179} & \underline{0.118}\\
    \bottomrule
\end{tabular}

    \caption{\textbf{Qualitative occlusion detection strategies.} Strategies during inference in the novel view. at $192\times 640$ resolution}
    \label{tab:occlusions_quantitative}
\end{table}
We measure the effect of the occlusion maps for different methods also quantitatively in \cref{tab:occlusions_quantitative}. 
As observed in the main paper, the depth gradient- based approach performs best.

\begin{figure}
    \centering
    \includegraphics[width=0.6\linewidth]{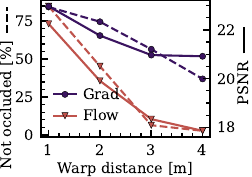}
    \caption{\textbf{VCM.} Gradient vs. flow-based occl. detection against warp distance.}
    \label{fig:sup_nvs_plot}
\end{figure}

\paragraph{Effect of Warp Distance and Occlusion Strategy on VCM.}
% The VCM works best if the occlusions are masked precisely and if there is sufficient scene context in the warped image.
As seen in \cref{fig:sup_nvs_plot}, our gradient-based occlusion detection strategy provides more precise masks ($\rightarrow$ more visual context) than the flow-based baseline.
Thus, the VCM produces better synthetic views.
When performing larger viewpoint changes, the quality degrades slightly.
In our experiments, novel view poses are on average $\sim3\operatorname{m}$ away from the source view(s).

\paragraph{Statistical Importance.} We report the mean and variance where possible in \cref{tab:reb_reconstruction_quantitative} \textit{a)}. The variance remains within limits, meaning that the final results are reliable.

\paragraph{Impact of supervision via the \pvolabbr.} To test the effect of our geometry synthesis independently of the VCM, we build the \pvol directly from real multi-view data (the same as in BTS) and report the results in \cref{tab:reb_reconstruction_quantitative} \textit{b)}.
While the resulting geometry is better than the one produced by BTS, it is still inferior to the geometry resulting from the synthetic views produced by the VCM.
This suggests that both the VCM and our novel way of obtaining scene geometry contribute to our strong performance.

\begin{table*}[t]
    \centering
    % \footnotesize
    % UniDepth & 89\textsubscript{\textit{0.84}} & -- & -- & 96\textsubscript{\textit{0.34}} & -- & -- \\
    % Metric3D & 91\textsubscript{\textit{0.64}} & -- & -- & 97\textsubscript{\textit{0.24}} & -- & -- \\
    % \textbf{Ours} (minor bug) & {93}\textsubscript{\textit{0.49}} & {72}\textsubscript{\textit{3.51}} & {74}\textsubscript{\textit{6.67}} & {97}\textsubscript{\textit{0.35}} & {73}\textsubscript{\textit{6.82}} & {96}\textsubscript{\textit{1.35}} \\
\setlength{\tabcolsep}{4pt} % Slightly reduced from the default of 6pt
\begin{tabular}{clccc|ccc}
    \toprule
    & \textit{Method} & $\text{O}_\text{acc}$~$\uparrow$ & $\text{IE}_\text{acc}$~$\uparrow$ & $\text{IE}_\text{rec}$~$\uparrow$ & $\text{O}_\text{acc}$~$\uparrow$ & $\text{IE}_\text{acc}$~$\uparrow$ & $\text{IE}_\text{rec}$~$\uparrow$ \\
    \midrule
    \multirow{4}{*}{\textit{a)}} 
    & BTS & 92\textsubscript{\textit{0.64}} & 69\textsubscript{\textit{4.24}} & 64\textsubscript{\textit{8.37}} & 95\textsubscript{\textit{0.74}} & 63\textsubscript{\textit{10.71}} & 94\textsubscript{\textit{1.88}} \\
    & BTS-D & 92\textsubscript{\textit{0.70}} & 70\textsubscript{\textit{3.96}} & 66\textsubscript{\textit{8.24}} & 95\textsubscript{\textit{0.77}} & 61\textsubscript{\textit{11.66}} & 96\textsubscript{\textit{1.11}} \\
    \cmidrule(lr){2-8}
    & \textbf{Ours} & 93\textsubscript{\textit{0.49}} & 72\textsubscript{\textit{3.51}} & 74\textsubscript{\textit{6.67}} 
        & 97\textsubscript{\textit{0.35}} 
        & 73\textsubscript{\textit{6.82}} 
        & 96\textsubscript{\textit{1.35}} \\
    & \textbf{Ours} (Distilled) & 90\textsubscript{\textit{0.86}} & 71\textsubscript{\textit{3.84}} & 71\textsubscript{\textit{9.00}} & 96\textsubscript{\textit{0.43}} & 72\textsubscript{\textit{7.10}} & 93\textsubscript{\textit{2.69}} \\
    \midrule
    \multirow{2}{*}{\textit{b)}} 
    & \textbf{Ours} & 93\textsubscript{\textit{0.49}} & 72\textsubscript{\textit{3.51}} & 74\textsubscript{\textit{6.67}} 
        & 97\textsubscript{\textit{0.35}} 
        & 73\textsubscript{\textit{6.82}} 
        & 96\textsubscript{\textit{1.35}} \\
    & Ours w/ real MV & 93\textsubscript{\textit{0.50}} & 71\textsubscript{\textit{3.91}} & 66\textsubscript{\textit{7.62}} & 97\textsubscript{\textit{0.25}} & 71\textsubscript{\textit{7.07}} & 91\textsubscript{\textit{3.25}} \\
    \bottomrule
\end{tabular}
    \caption{\textbf{Scene Reconstruction.} \textit{a)} Results on KITTI-360 / Waymo in \% with variances. \textit{b)} Comparing synthesized views vs. real multi-view data for building the \pvol.
    }
    \label{tab:reb_reconstruction_quantitative}
\end{table*}

\textbf{Inference Latency.}
% Inference time, Comparability, Consistency
While the \pvolabbr geometry is of high-quality, the generation process is costly at $5303\frac{\operatorname{ms}}{\operatorname{frame}}$ on an A100 GPU. 
The distilled \recon only takes about $75\frac{\operatorname{ms}}{\operatorname{frame}}$ ($>$70x faster) and is real-time capable.
It also produces more stable results, while the original synthesized geometry sometimes contains artifacts.
Both aspects are crucial for the applications we target (autonomous driving, robotics, etc.).